\documentclass[journal]{IEEEtran}
\IEEEoverridecommandlockouts
\usepackage{tikz}
\usepackage{comment}
\usepackage{pgfplots}
\usepackage[
    colorlinks=false,   
    pdfborder={0 0 1},  
    linkbordercolor={0 1 0},
    citebordercolor={0 1 0},
    urlbordercolor={0 1 0}
]{hyperref}
\pgfplotsset{compat=1.18}
\usepackage{xcolor}
\usepackage{graphicx}
\usepackage{subcaption}
\providecolor{latencyorange}{RGB}{215,116,48}
\providecolor{efficiencygreen}{RGB}{39,148,113}
\providecolor{tokenpurple}{RGB}{145,111,194}
\usepackage{booktabs}
\usepackage{pifont}
\usepackage{makecell}

\usepackage{tikz}
\usepackage{pgfplots}
\usepackage{xcolor}

\pgfplotsset{compat=1.18}

\definecolor{leaderboardblue}{RGB}{156,204,235}

\pgfplotsset{
  leaderboardaxis/.style={
    width=\linewidth,
    height=9.2cm,
    xmin=0,
    xmax=90,
    xtick={0,20,40,60,80},
    ymin=-0.55,
    ymax=14.55,
    y dir=reverse,
    ytick={0,...,14},
    xlabel={Accuracy (\%)},
    xlabel style={font=\scriptsize},
    xticklabel style={font=\tiny},
    yticklabel style={
      font=\fontsize{6}{6.6}\selectfont,
      align=right,
      text width=3.15cm,
      xshift=-2pt
    },
    ytick style={draw=none},
    xmajorgrids,
    grid style={black!12},
    axis x line*=bottom,
    axis y line*=left,
    axis line style={
      draw=black!55,
      line width=0.3pt
    },
    clip=false
  }
}

\pgfplotsset{
  leaderboardbar/.style={
    xbar,
    bar width=7.5pt,
    bar shift=0pt,
    fill=leaderboardblue,
    draw=none,
    error bars/x dir=both,
    error bars/x explicit,
    error bars/error bar style={
      draw=black,
      line width=0.45pt
    },
    error bars/error mark options={
      draw=black,
      rotate=90,
      mark size=1.8pt,
      line width=0.45pt
    }
  }
}

\newcommand{\leadercfg}[2]{%
  \node[
    anchor=west,
    font=\tiny,
    text=black
  ] at (axis cs:1,#1) {#2};
}

\newcommand{\leadermean}[4]{%
  \pgfmathsetmacro{\leaderlabelx}{#2+#3+0.7}%
  \node[
    anchor=west,
    font=\tiny,
    text=black
  ] at (axis cs:\leaderlabelx,#1) {#4};
}

\newcommand{\cmark}{\ding{51}}
\newcommand{\xmark}{\ding{55}}
\newcommand{\pmark}{$\bullet$}   

\usepackage{tikz}
\usepackage{pgfplots}
\usepackage{xcolor}
\usepackage{graphicx}

\pgfplotsset{compat=1.18}

\definecolor{promptblue}{RGB}{200,75,75}

\pgfplotsset{
  promptaxis/.style={
    width=\linewidth,
    height=9.2cm,
    xmin=0,
    xmax=96,
    xtick={0,20,40,60},
    xlabel={Accuracy (\%)},
    xlabel style={font=\scriptsize},
    xticklabel style={font=\tiny},
    y dir=reverse,
    yticklabel style={
      font=\fontsize{6}{6.6}\selectfont,
      align=right,
      text width=3.15cm,
      xshift=-2pt
    },
    ytick style={draw=none},
    xmajorgrids,
    grid style={black!10},
    axis x line*=bottom,
    axis y line*=left,
    axis line style={
      draw=black!55,
      line width=0.3pt
    },
    clip=false
  }
}

\newcommand{\promptconnect}[4]{%
  \draw[
    black!35,
    line width=0.45pt
  ]
  (axis cs:#2,#1) --
  (axis cs:#3,#1) --
  (axis cs:#4,#1);
}

\usepackage{tikz}
\usepackage{xcolor}

\definecolor{latencyorange}{RGB}{215,120,55}
\definecolor{efficiencygreen}{RGB}{35,145,115}
\definecolor{tokenpurple}{RGB}{125,90,180}

\definecolor{temperatureteal}{RGB}{35,145,135}

\newcommand{\temperaturecell}[3]{%
  \pgfmathtruncatemacro{\tempshade}
  {max(12,min(72,12+60*(#3-20)/45))}%
  \node[
    minimum width=0.88cm,
    minimum height=0.42cm,
    inner sep=0pt,
    draw=white,
    line width=0.25pt,
    fill=temperatureteal!\tempshade!white,
    font=\tiny
  ] at (#1,#2) {%
    \ifdim #3pt>50pt
      \color{white}%
    \else
      \color{black}%
    \fi
    #3%
  };
}

\newcommand{\temperaturestd}[3]{%
  \pgfmathtruncatemacro{\stdshade}
  {max(8,min(75,8+67*#3/8.5))}%
  \node[
    minimum width=0.78cm,
    minimum height=0.42cm,
    inner sep=0pt,
    draw=white,
    line width=0.25pt,
    fill=black!\stdshade,
    font=\tiny
  ] at (#1,#2) {%
    \ifdim #3pt>4pt
      \color{white}%
    \else
      \color{black}%
    \fi
    #3%
  };
}

\newcommand{\temperatureheader}{%
  \node[
    anchor=east,
    font=\scriptsize\bfseries
  ] at (-0.03,0.50) {Model};
  \node[font=\scriptsize\bfseries] at (0.50,0.50) {$T=0.0$};
  \node[font=\scriptsize\bfseries] at (1.45,0.50) {$T=0.2$};
  \node[font=\scriptsize\bfseries] at (2.40,0.50) {$T=0.5$};
  \node[font=\scriptsize\bfseries] at (3.35,0.50) {$T=0.8$};
  \node[font=\scriptsize\bfseries] at (4.35,0.50) {Std};
}

\definecolor{robustpurple}{RGB}{125,90,180}

\pgfplotsset{
  robustaxis/.style={
    width=\linewidth,
    height=9.2cm,
    xmin=0,
    xmax=100,
    xtick={0,20,40,60},
    xlabel={Accuracy (\%)},
    xlabel style={font=\scriptsize},
    xticklabel style={font=\tiny},
    y dir=reverse,
    yticklabel style={
      font=\fontsize{5.7}{6.2}\selectfont,
      align=right,
      text width=3.15cm,
      xshift=-2pt
    },
    ytick style={draw=none},
    xmajorgrids,
    grid style={black!10},
    axis x line*=bottom,
    axis y line*=left,
    axis line style={
      draw=black!55,
      line width=0.3pt
    },
    clip=false
  }
}

\newcommand{\robustrow}[5]{%
  \pgfmathsetmacro{\robustbest}{#3+#4}%
  \draw[
    robustpurple!40,
    line width=2.6pt,
    line cap=round
  ]
  (axis cs:#3,#1) --
  (axis cs:\robustbest,#1);
  \node[
    circle,
    draw=robustpurple,
    fill=white,
    line width=0.6pt,
    minimum size=3.8pt,
    inner sep=0pt
  ] at (axis cs:#3,#1) {};
  \node[
    circle,
    draw=robustpurple,
    fill=robustpurple,
    line width=0.5pt,
    minimum size=3.8pt,
    inner sep=0pt
  ] at (axis cs:\robustbest,#1) {};
  \node[
    rectangle,
    rotate=45,
    draw=robustpurple!80!black,
    fill=robustpurple!80!black,
    line width=0.4pt,
    minimum size=4.4pt,
    inner sep=0pt
  ] at (axis cs:#2,#1) {};
  \node[
    anchor=south,
    yshift=1.5pt,
    font=\fontsize{4.7}{5}\selectfont
  ] at (axis cs:#2,#1) {#2};
  \node[
    anchor=north,
    yshift=-1.4pt,
    font=\fontsize{4.7}{5}\selectfont
  ] at (axis cs:#3,#1) {#3};
  \node[
    anchor=west,
    font=\fontsize{5}{5.4}\selectfont
  ] at (axis cs:77,#1) {#4\,/\,#5};
}
\newcommand{\temperaturerow}[7]{%
  \pgfmathsetmacro{\tempy}{-0.46*#1}%
  \node[
    anchor=east,
    font=\fontsize{5.5}{6.0}\selectfont,
    text width=3.15cm,
    align=right,
    inner sep=0pt
  ] at (-0.03,\tempy) {#2};
  \temperaturecell{0.50}{\tempy}{#3}
  \temperaturecell{1.45}{\tempy}{#4}
  \temperaturecell{2.40}{\tempy}{#5}
  \temperaturecell{3.35}{\tempy}{#6}
  \temperaturestd{4.35}{\tempy}{#7}
}

\newcommand{\promptgain}[2]{%
  \node[
    anchor=west,
    font=\tiny,
    text=black
  ] at (axis cs:78,#1) {#2};
}
\usepackage{booktabs}
\usepackage{pifont}
 \usepackage{url}
 \usepackage[most]{tcolorbox}

\newtcolorbox{rqbox1}{
    colback=gray!8,
    colframe=black,
    boxrule=0.8pt,
    arc=2pt,
    left=6pt,
    right=6pt,
    top=6pt,
    bottom=6pt,
    title=System Prompt,
    fonttitle=\bfseries
}
\definecolor{GOne}{RGB}{33,113,181}     
\definecolor{GTwo}{RGB}{35,139,69}      
\definecolor{GThree}{RGB}{146,36,40}    
\definecolor{GFour}{RGB}{117,107,177}   
\definecolor{GFive}{RGB}{217,95,2}      
\definecolor{EdgeColor}{RGB}{0,90,120}      
\definecolor{EdgeColorLight}{RGB}{0,130,160}
\usepackage{booktabs}
\usepackage{cite}
\usepackage{amsmath,amssymb,amsfonts}
\usepackage{algorithm}
\usepackage{algpseudocode}
\usepackage{graphicx}
\usepackage{textcomp}
\usepackage{xcolor}
\def\BibTeX{{\rm B\kern-.05em{\sc i\kern-.025em b}\kern-.08em
    T\kern-.1667em\lower.7ex\hbox{E}\kern-.125emX}}
\begin{document}

\title{\textcolor{black}{PhysAI-Bench: A Benchmark for LLM-Based Agentic Decision-Making in Autonomous UAV-Centric Physical AI}}
\author{
\IEEEauthorblockN{
Mohamed~Amine~Ferrag\IEEEauthorrefmark{1}\IEEEauthorrefmark{5},
Merouane~Debbah\IEEEauthorrefmark{2}, Abderrahmane~Lakas\IEEEauthorrefmark{1}, Manu Perumkunnil\IEEEauthorrefmark{3}, Norbert Tihanyi \IEEEauthorrefmark{4}
}
\\
\IEEEauthorblockA{\IEEEauthorrefmark{1}
College of Computing and Artificial Intelligence (CCAI),  
United Arab Emirates University, UAE
} \\
\IEEEauthorblockA{\IEEEauthorrefmark{2}
Research Institute for Digital Future, Khalifa University, UAE
}\\
\IEEEauthorblockA{\IEEEauthorrefmark{3}
Interuniversity Microelectronics Centre (IMEC), Belgium
}\\
\IEEEauthorblockA{\IEEEauthorrefmark{4}
Technology Innovation Institute, Abu Dhabi, UAE
}\\
\IEEEauthorblockA{\IEEEauthorrefmark{5}
Corresponding author: \texttt{mohamed.ferrag@uaeu.ac.ae}
}

\thanks{This work was supported by United Arab Emirates University under the Research Start-up Program (Grant No. G00005769).}

}

\maketitle

\begin{abstract}
Recent advances in Physical AI have accelerated the deployment of foundation models in autonomous systems, particularly unmanned aerial vehicles (UAVs), that continuously perceive, reason, plan, and interact with dynamic physical environments. Although existing benchmarks evaluate complementary capabilities such as physical perception, intuitive physics, embodied navigation, and collaborative reasoning, they rarely assess the underlying agentic decision-making process required for reliable autonomous operation. To address this limitation, we present \textit{PhysAI-Bench}, a benchmark for evaluating agentic decision-making in Physical AI systems. In its current instantiation, \textit{PhysAI-Bench} comprises 10{,}178 standardized decision-making instances automatically constructed from large-scale conversational traces of autonomous UAV missions by extracting trace-derived decision points. Each instance preserves mission context, temporal dependencies, operational semantics, physical constraints, Model Context Protocol (MCP) tool invocations, Agent-to-Agent (A2A) interactions, sensor observations, and AI-native 6G networking conditions, including latency, packet loss, throughput, edge-computing load, and network slicing. Only information available before the selected decision is exposed, preventing direct leakage from subsequent trace events and approximating online autonomous decision-making. We evaluate 29 foundation models using a two-stage protocol. First, we evaluate 12 combinations of zero-shot, three-shot, and five-shot prompting and four decoding temperatures in three repeated runs on a 35-instance human-verified development set. The model-specific configuration selected at this stage is then frozen and evaluated in three prespecified runs on a fixed, episode-disjoint held-out set of 500 decision instances. On the held-out set, GPT-5.3 achieves the highest decision accuracy (52.00\%), followed by GPT-5.2 (49.40\%) and Grok~4.5 (49.07\%). Development-stage analyses further show that few-shot prompting generally improves performance for most models, whereas decoding temperature has comparatively limited influence. These results show that reliable agentic decision-making under Physical AI conditions remains a challenging open problem. To support open science and reproducibility, we release the \textit{PhysAI-Bench} dataset on GitHub: \url{https://github.com/maferrag/physai-bench}.\end{abstract}


\section{Introduction}

Recent advances in Large Language Models (LLMs) have transformed artificial intelligence from task-specific learning systems into increasingly general-purpose reasoning engines that can understand natural language, perform complex multi-step reasoning, and interact with external tools \cite{bajoria2026survey, FERRAG20251054}. Building on these capabilities, Agentic AI has further extended foundation models with autonomous planning, memory, tool use, and long-horizon decision-making \cite{javaid2024large}. Meanwhile, emerging interoperability frameworks, such as the MCP, which provides standardized access to external tools and knowledge sources, and A2A communication, which enables collaboration among autonomous agents, are accelerating the shift from passive conversational assistants to intelligent systems that can operate autonomously in complex real-world environments \cite{ferrag2026llm}.

These developments are driving the rapid emergence of \emph{Physical AI}, where intelligent agents continuously perceive dynamic environments, reason over multimodal observations, plan under uncertainty, and execute actions that directly influence the physical world \cite{team2025gemini, krojer2025shortcut}. Unlike conventional language-centric AI systems that primarily generate textual responses, Physical AI agents operate within closed perception--reasoning--action loops while interacting with sensors, actuators, external tools, and collaborating agents \cite{xue2026acwm}. Such capabilities are becoming increasingly important across robotics, autonomous driving, unmanned aerial vehicles (UAVs), embodied assistants, industrial automation, smart manufacturing, and other cyber-physical systems. Recent foundation models, such as NVIDIA's Cosmos~3 \cite{agarwal2026cosmos}, further demonstrate this evolution by jointly modeling language, vision \cite{motamed2026generative}, video, audio, and actions within a unified world model capable of multimodal understanding, world simulation, and policy generation, highlighting the growing convergence between foundation models and autonomous Physical AI systems \cite{zhou2026chatvla}.

The rapid evolution of Physical AI has also stimulated significant progress in evaluation methodologies. On one hand, recent benchmark datasets, including PAI-Bench \cite{zhou2025paibenchcomprehensivebenchmarkphysical}, Causalvqa \cite{foss2025causalvqa}, T2vphysbench \cite{guo2025t2vphysbench}, IntPhys~2 \cite{bordes2025intphys}, DeepPHY \cite{xu2026deepphy}, PHYSGYM \cite{chen2026physgym}, LMEE-Bench \cite{wang2026explore}, NavBench \cite{qiao2026navbench}, AirCopBench \cite{zha2026aircopbench}, and $\alpha^3$-Bench \cite{11609268}, evaluate complementary capabilities such as physical perception, intuitive physics, embodied navigation, interactive reasoning, collaborative perception, and autonomous agent behaviour. On the other hand, state-of-the-art Physical AI foundation models are increasingly evaluated using collections of specialized benchmark suites, such as PAIBench-G \cite{zhou2025paibenchcomprehensivebenchmarkphysical}, RBench \cite{deng2026rethinking}, Physics-IQ \cite{motamed2026generative}, Cosmos-HUE \cite{nvidia2026cosmoshumaneval}, Human World Bench (HWB), RoboArena \cite{atreya2025roboarena}, and RoboLab \cite{yang2026robolab}, reflecting the multidimensional nature of Physical AI evaluation. Although these benchmarks have substantially advanced the evaluation of individual capabilities, they primarily assess perception quality, world understanding, simulation fidelity, embodied interaction, or task completion rather than the underlying agentic decision-making process itself.

Evaluating agentic decision-making presents unique challenges beyond conventional language understanding or perception benchmarks \cite{zaidi2026reasoning}. Autonomous agents must continuously select appropriate actions from incomplete observations while considering mission objectives, environmental dynamics, physical constraints, external tool responses, and interactions with collaborating agents \cite{chergui2026tutorial}. Consequently, measuring only reasoning accuracy or task completion provides an incomplete assessment of the capabilities required for reliable autonomous operation \cite{jiang2026robowm}. Furthermore, existing benchmarks rarely evaluate decision-making using realistic autonomous execution traces while explicitly preventing future information leakage, making it difficult to quantify the true decision-making capability, robustness, and generalization of modern Physical AI foundation models \cite{chen2026abot,cho2026spatialclaw}.

\textcolor{black}{%
To address these limitations, we propose \textit{PhysAI-Bench}, a benchmark that converts autonomous UAV mission traces into multiple-choice action-selection tasks.%
}
The main contributions are as follows:

\begin{itemize}

    \item \textbf{Benchmark dataset:} \textit{PhysAI-Bench} contains \(10{,}178\) trace-derived decision instances covering partial observability, changing mission objectives, environmental uncertainty, physical constraints, sensor feedback, and AI-native 6G networking conditions.

    \item \textbf{Automated construction pipeline:} A schema-driven pipeline extracts UAV decision points, removes subsequent trace information, and preserves the mission context, temporal dependencies, MCP tool invocations, A2A interactions, sensor observations, and network conditions available at decision time.

    \item \textbf{Large-scale evaluation:} We evaluate 29 frontier LLMs from \(14\) AI organizations using a two-stage protocol. We first test twelve prompting--temperature configurations over three runs on a \(35\)-instance human-verified development set. We then freeze the selected configuration for each model and evaluate it over three runs on an episode-disjoint held-out set of \(500\) instances. GPT-5.3 Chat achieves the highest held-out accuracy of \(52.00\%\), followed by GPT-5.2 Chat with \(49.40\%\) and Grok~4.5 with \(49.07\%\). Few-shot prompting generally improves development-stage performance, whereas temperature has limited influence.

\end{itemize}

The remainder of this paper is organized as follows. Section \ref{sec:sec2} reviews the literature on Physical AI, embodied intelligence, and agentic AI benchmarks, and positions \textit{PhysAI-Bench} relative to current evaluation methodologies. Section \ref{sec:sec3} presents the proposed \textit{PhysAI-Bench} framework, including its design objectives, formal benchmark formulation, benchmark construction methodology, automated generation pipeline, and comprehensive statistical characterization. Section \ref{sec:sec4} describes the experimental methodology, including the evaluated large language models, prompting strategies, inference settings, and evaluation metrics. Section \ref{sec:experimental_Results} presents and analyzes the experimental results, including overall benchmark performance, the impact of prompting strategies and decoding temperatures, robustness across inference configurations, and computational efficiency. Finally, Section \ref{sec:conc} concludes the paper and discusses the main findings, limitations, and future research directions.

\begin{table*}[!t]
\centering
\caption{Comparison of representative Physical AI and embodied-agent benchmarks.}
\label{tab:benchmark_comparison}

\footnotesize
\setlength{\tabcolsep}{4pt}

\begin{tabular}{lccccccccc}
\toprule
\textbf{Benchmark} &
\textbf{Year} &
\makecell{\textbf{Physical}\\\textbf{AI}} &
\makecell{\textbf{Autonomous}\\\textbf{Agents}} &
\makecell{\textbf{Autonomous}\\\textbf{Decision-Making}} &
\makecell{\textbf{Interactive}\\\textbf{Reasoning}} &
\makecell{\textbf{Physical}\\\textbf{Context}} &
\makecell{\textbf{6G Network}\\\textbf{Context}} &
\makecell{\textbf{MCP}\\\textbf{/ A2A}} \\
\midrule

PAI-Bench~\cite{zhou2025paibenchcomprehensivebenchmarkphysical}
& 2025
& \cmark & \xmark & $\times$ & \pmark & \cmark & \xmark & \xmark \\

PBench~\cite{nvidia2025pbench}
& 2025
& $\checkmark$
& $\times$
& $\times$
& $\times$
& $\checkmark$
& $\times$
& $\times$ \\

IntPhys~2~\cite{bordes2025intphys}
& 2025
& \cmark & \xmark & $\times$ & \xmark & \cmark & \xmark & \xmark \\

DeepPHY~\cite{xu2026deepphy}
& 2026
& \cmark & \pmark & $\times$ & \cmark & \cmark & \xmark & \xmark \\

PHYSGYM~\cite{chen2026physgym}
& 2026
& \cmark & \pmark & $\times$ & \cmark & \cmark & \xmark & \xmark \\

LMEE-Bench~\cite{wang2026explore}
& 2026
& \pmark & \cmark & $\times$ & \cmark & \pmark & \xmark & \xmark \\

$\alpha^3$-Bench~\cite{11609268}
& 2026
& $\times$ & \cmark & $\times$ & \cmark & \cmark & \cmark & \cmark \\

NavBench~\cite{qiao2026navbench}
& 2026
& \pmark & \cmark & $\times$ & \cmark & \pmark & \xmark & \xmark \\

AirCopBench~\cite{zha2026aircopbench}
& 2026
& \cmark & \cmark & $\times$ & \cmark & \cmark & \xmark & \pmark \\

\midrule

\textbf{PhysAI-Bench}
& \textbf{2026}
& \cmark & \cmark & \cmark & \cmark & \cmark & \cmark & \cmark \\

\bottomrule
\end{tabular}

\vspace{2mm}
\footnotesize{$\checkmark$: supported; $\bullet$: partially supported; $\times$: not supported.}
\end{table*}

\section{Related Work}
\label{sec:sec2}
The rapid emergence of Physical AI has spurred the development of numerous benchmark datasets to evaluate foundation models across diverse capabilities, including physical perception, intuitive physics, embodied reasoning, autonomous navigation, collaborative intelligence, and agentic decision-making. To position \textit{PhysAI-Bench} within this evolving landscape, we organize the existing literature into two complementary categories. We first review representative benchmarks for Physical AI and physical reasoning, followed by benchmarks targeting embodied and autonomous agents. Finally, we discuss how \textit{PhysAI-Bench} positions itself relative to existing evaluation methodologies and highlight its unique contributions.

\subsection{Physical AI and Physical Reasoning Benchmarks}

Zhou et al.~\cite{zhou2025paibenchcomprehensivebenchmarkphysical} proposed PAI-Bench, a comprehensive benchmark for evaluating Physical AI from the perspectives of perception and prediction using videos. The benchmark comprises three complementary tracks covering video generation, conditional video generation, and video understanding, with a total of 2,808 real-world evaluation instances spanning autonomous driving, robotics, industry, egocentric scenarios, human activities, and physical commonsense reasoning.  Distinct from PAI-Bench, NVIDIA’s PBench \cite{nvidia2025pbench} evaluates whether world models can generate physically and semantically consistent future videos from an initial image and a textual prompt. The benchmark contains 1,044 samples and 5,636 binary question–answer pairs spanning autonomous driving, robotics, industrial environments, physics, human activities, and common-sense scenarios.

Bordes et al.~\cite{bordes2025intphys} proposed IntPhys~2, a benchmark for evaluating intuitive physics understanding in complex synthetic environments. Building upon the original IntPhys benchmark, IntPhys~2 assesses four fundamental physical principles, namely object permanence, immutability, spatio-temporal continuity, and solidity, using a violation-of-expectation framework that requires models to distinguish physically plausible from implausible events. The authors evaluated several state-of-the-art multimodal large language models and predictive world models, reporting that most systems performed near random chance (approximately 50\% accuracy), whereas human annotators achieved about 96\% accuracy, showing that intuitive physics reasoning remains a significant challenge for current AI models. Xu et al.~\cite{xu2026deepphy} proposed DeepPHY, a benchmark for evaluating the interactive physical reasoning capabilities of agentic Vision--Language Models (VLMs) in dynamic simulated environments. Unlike static physics question-answering benchmarks, DeepPHY evaluates models through continuous interaction across six physics-based environments, namely PHYRE, I-PHYRE, Kinetix, Pooltool, Angry Birds, and Cut the Rope, covering fundamental physics, sequential planning, and action control. Chen et al.~\cite{chen2026physgym} proposed PHYSGYM, an interactive benchmark for evaluating the scientific reasoning capabilities of LLM-based agents in physics discovery tasks. Unlike static equation-discovery benchmarks, PHYSGYM formulates scientific discovery as an interactive process in which agents iteratively design experiments, collect observations under a limited experimental budget, and infer the underlying physical equations.

\subsection{Embodied and Autonomous Agent Benchmarks}

Ferrag et al.~\cite{11609268} proposed $\alpha^3$-Bench, a benchmark for evaluating LLM-based autonomous UAV agents through multi-turn conversational reasoning under dynamic 6G network conditions. The benchmark formulates UAV missions as language-mediated control loops that integrate MCP tool calls, A2A coordination, and network-aware reasoning across varying latency, packet loss, throughput, and edge-computing conditions. It is constructed from more than 113,000 AI conversational UAV mission episodes derived from UAVBench scenarios and evaluates 17 state-of-the-art LLMs using a fixed subset of 50 episodes per scenario. The authors further introduce the composite $\alpha^3$ metric, which jointly measures task outcome, safety policy compliance, tool consistency, interaction quality, network robustness, and communication cost. Wang et al.~\cite{wang2026explore} proposed LMEE-Bench, a benchmark for evaluating long-term memory embodied exploration by jointly assessing multi-goal navigation and memory-based question answering. The benchmark introduces a new evaluation paradigm in which agents must continuously collect episodic memories during exploration and later retrieve them to answer questions about previously observed environments. LMEE-Bench comprises 208 training scenes and 38 test scenes, including 8,880 training goals, 828 test goals, and 9,286 generated questions spanning five memory-reasoning categories.  Qiao et al.~\cite{qiao2026navbench} proposed NavBench, a benchmark to evaluate the embodied navigation capabilities of multimodal large language models in zero-shot settings. The benchmark decomposes navigation into two complementary components: navigation comprehension and navigation execution. Navigation comprehension is assessed through three cognitively grounded tasks, namely global instruction alignment, temporal progress estimation, and local observation--action reasoning, comprising 3,200 question--answer pairs, while navigation execution evaluates step-by-step decision-making over 432 navigation episodes across 72 indoor scenes with three difficulty levels based on spatial, cognitive, and execution complexity. Zha et al.~\cite{zha2026aircopbench} proposed AirCopBench, a benchmark for evaluating multimodal large language models in collaborative multi-UAV embodied perception and reasoning under challenging real-world conditions. AirCopBench comprises more than 14.6k visual question-answer pairs generated from 2,920 synchronized multi-view images collected from both simulated and real-world environments, covering 14 task types grouped into four evaluation dimensions: scene understanding, object understanding, perception assessment, and collaborative decision-making.

\subsection{Positioning of PhysAI-Bench}

Table~\ref{tab:benchmark_comparison} compares \textit{PhysAI-Bench} with representative Physical AI and embodied-agent benchmarks, highlighting its distinctive and complementary role within the current evaluation landscape. Existing benchmarks assess capabilities such as physical perception, intuitive physics, scientific reasoning, embodied navigation, long-term memory, collaborative perception, and conversational autonomous agents~\cite{zhou2025paibenchcomprehensivebenchmarkphysical,bordes2025intphys,xu2026deepphy,chen2026physgym,wang2026explore,qiao2026navbench,zha2026aircopbench,11609268}. However, these benchmarks mainly evaluate perception quality, physical understanding, navigation success, task completion, collaborative reasoning, or end-to-end autonomous execution rather than the underlying decision-making process. In particular, $\alpha^3$-Bench~\cite{11609268} evaluates complete conversational UAV missions under dynamic 6G network conditions using composite metrics for task success, safety, interaction quality, network robustness, and communication efficiency. Although these metrics provide a comprehensive view of mission performance, they can obscure inappropriate intermediate decisions later corrected through replanning, as well as appropriate decisions followed by mission failure due to tool errors, communication impairments, or adverse environmental conditions. In contrast, \textit{PhysAI-Bench} evaluates individual decision points extracted from autonomous UAV mission traces and exposes only the information available before each action, thereby preventing future-information leakage and separating action-selection capability from downstream execution reliability. By preserving mission context, physical constraints, sensor observations, MCP tool invocations, A2A interactions, and network conditions, \textit{PhysAI-Bench} enables fine-grained analysis of navigation, safety, sensing, coordination, and network-aware reasoning while providing a scalable and reproducible framework for evaluating agentic decision-making in Physical AI.

\begin{figure*}[t]
    \centering
    \includegraphics[width=\textwidth]
    {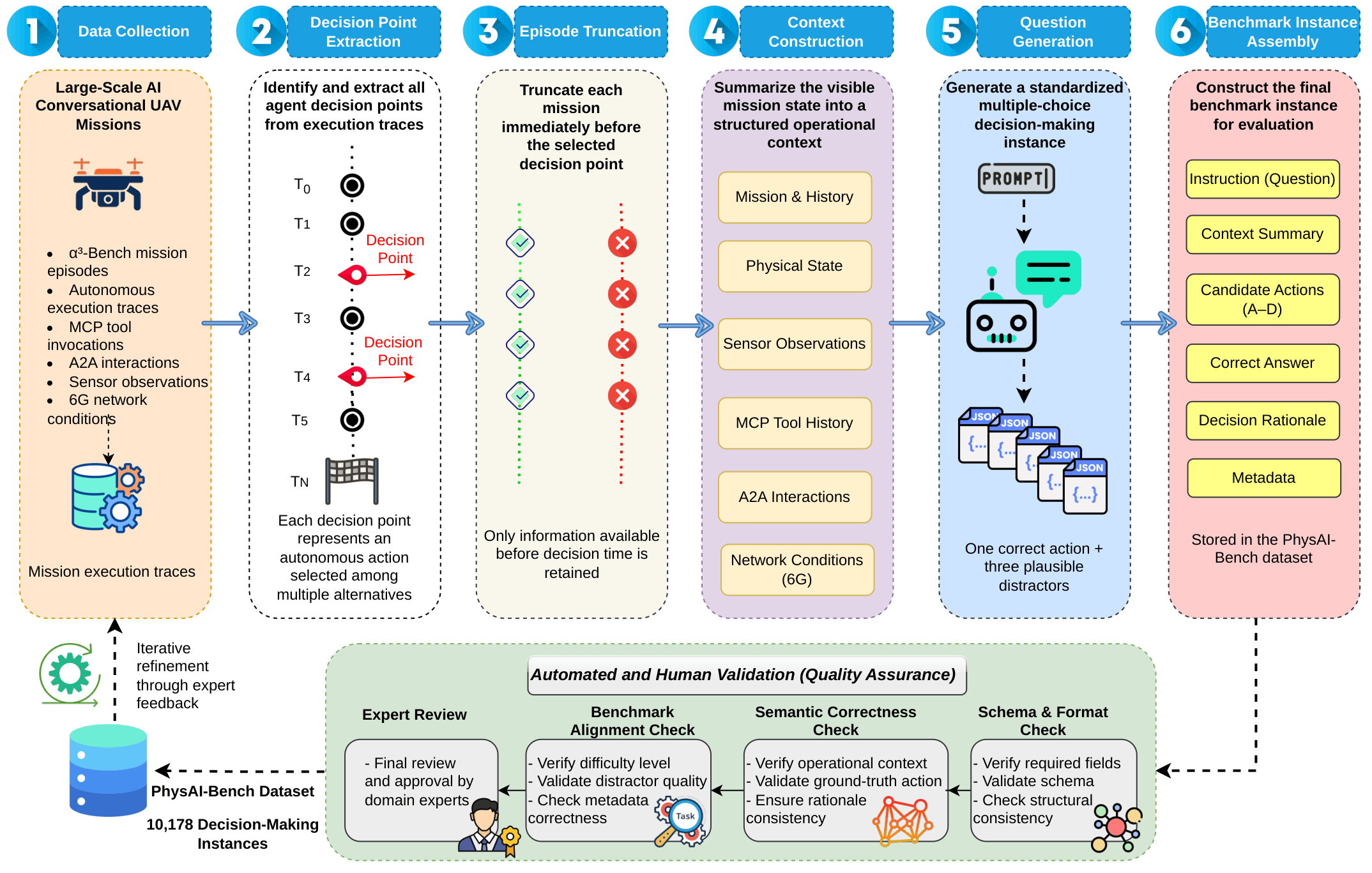}
    \caption{End-to-end construction and validation workflow for \textit{PhysAI-Bench}. Conversational UAV mission traces are converted into multiple-choice decision instances by identifying decision points, truncating pre-decision context, constructing context, generating questions, validating automatically, and expert review.}
    \label{fig:physai_generation_pipeline}
\end{figure*}

\section{PhysAI-Bench}
\label{sec:sec3}

In this section, we introduce PhysAI-Bench, a benchmark specifically designed to evaluate the decision-making capabilities of foundation models in Physical AI. Unlike existing benchmarks that primarily assess perception, language understanding, or task completion, \textit{PhysAI-Bench} focuses on selecting the most appropriate action under realistic physical and operational constraints. We first present the benchmark design objectives, followed by its formal formulation, benchmark construction methodology, and automated generation pipeline, before summarizing the resulting benchmark statistics.

\subsection{Design Objectives}

\textit{PhysAI-Bench} is motivated by the rapid emergence of Physical AI, where intelligent agents are expected to perceive, reason, plan, and interact with the physical world through autonomous decision-making. Recent benchmarks, such as $\alpha^3$-Bench \cite{11609268}, formulate autonomous UAV missions as multi-turn AI conversational reasoning problems that integrate MCP tool invocation, A2A communication, and dynamic 6G networking to evaluate conversational autonomous agents. While these benchmarks provide valuable insights into conversational reasoning, protocol compliance, and agent collaboration, they primarily assess dialogue-driven interactions within a specific application domain. In contrast, \textit{PhysAI-Bench} evaluates the fundamental decision-making capabilities that underpin Physical AI systems, where an intelligent agent must infer the most appropriate action from its current perception of the environment while accounting for mission objectives, environmental dynamics, and physical constraints.

To achieve this objective, \textit{PhysAI-Bench} is constructed around several key design principles. First, we derive every benchmark instance from an actual agent decision point extracted from autonomous execution traces, ensuring each question reflects a realistic decision encountered during task execution. Second, we expose only the information available before the selected decision to the evaluated model, preventing future information leakage and faithfully simulating online decision-making. Third, benchmark instances are formulated as structured multiple-choice action-selection problems that enable objective, scalable, and fully automated evaluation while preserving the semantics of real agent behaviours. Finally, the benchmark generation pipeline adopts a fully automated, schema-driven, and reproducible construction process, extending beyond AI conversational UAV missions to support a broader spectrum of Physical AI applications, including robotics, autonomous driving, embodied AI, UAVs, industrial automation, and other cyber-physical systems.

\subsection{Benchmark Overview}

Fig.~\ref{fig:physai_generation_pipeline} illustrates the complete
\textit{PhysAI-Bench} construction pipeline, including decision-point extraction,
future-information removal, context construction, benchmark-instance
generation, and automated and human quality assurance.

\subsubsection{Physical AI Environment}

\textit{PhysAI-Bench} models a Physical AI system as an autonomous agent interacting with a dynamic and partially observable physical environment. At each decision step, the agent continuously perceives its surroundings, reasons over available contextual information, and selects an action that influences the environment's subsequent evolution. Formally, the environment is represented as
\begin{equation}
\mathcal{E} = (\mathcal{S}, \mathcal{O}, \mathcal{A}, \mathcal{T}),
\label{eq:environment}
\end{equation}
where $\mathcal{S}$ denotes the environment state space, $\mathcal{O}$ the observation space available to the agent, $\mathcal{A}$ the admissible action space, and $\mathcal{T}(s_{t+1}\mid s_t,a_t)$ the stochastic transition function describing the probability of reaching state $s_{t+1}$ after executing action $a_t$ in state $s_t$. In PhysAI-Bench, environment evolution is captured through the AI conversational episodes generated by the $\alpha^3$-Bench framework, where state transitions arise from interactions between the autonomous agent, the environment, and external entities via structured dialogues, MCP tool invocations, A2A communication, and dynamic network conditions. Unlike traditional language-centric benchmarks, \textit{PhysAI-Bench} evaluates reasoning in environments where actions directly modify the system's physical state and future observations depend on both the agent's decisions and the underlying environment dynamics.

Since autonomous agents rarely have access to the complete environment state, benchmark instances are constructed from partial observations rather than full state information. Let $s_t\in\mathcal{S}$ denote the environment state at decision step $t$. The information available to the agent is defined as
\begin{equation}
o_t = f(s_t,\eta_t), \qquad o_t \in \mathcal{O},
\label{eq:observation}
\end{equation}
where $f(\cdot)$ denotes the observation function and $\eta_t$ models uncertainty arising from sensing, communication, and environmental conditions. In the generated AI conversational episodes, observations may include physical state variables, sensor outputs, network status, MCP tool responses, and A2A coordination messages, depending on the current mission context. Consequently, every benchmark instance evaluates the model under realistic partial observability, requiring it to infer the most appropriate action solely from the information available before the decision while preventing any access to future states or actions.

\subsubsection{Agent Decision Process}

Within each AI conversational episode, the Physical AI agent performs sequential decision-making by reasoning over the information accumulated throughout the mission. Rather than relying solely on the current observation, each decision conditions on the conversational context, including previous observations, user instructions, MCP tool responses, A2A coordination messages, and the current environment state. Let
\begin{equation}
h_t=\left\{d_1,d_2,\ldots,d_{t-1},o_t\right\}
\label{eq:history}
\end{equation}

denotes the decision context available immediately before decision step $t$, where $d_i$ represents the preceding dialogue turns, and $o_t$ is the current observation. The agent is modeled as a decision policy
\begin{equation}
\pi_{\theta}: \mathcal{H}\rightarrow\mathcal{A},
\label{eq:policy}
\end{equation}
where $\mathcal{H}$ denotes the space of conversational decision contexts and $\theta$ represents the parameters of the evaluated foundation model. The predicted action is therefore given by
\begin{equation}
a_t=\pi_{\theta}(h_t), \qquad a_t\in\mathcal{A},
\label{eq:decision}
\end{equation}
where $a_t$ denotes the action the model selects.

Unlike conventional language-generation tasks, the objective is not to produce free-form textual responses but to infer the action that best satisfies the operational objectives, environmental conditions, and physical constraints encoded in the conversational context. During benchmark generation, the complete decision context $h_t$ is transformed into a structured context summary that preserves only the information available before the selected decision, while excluding the executed action and all subsequent dialogue turns. Consequently, each benchmark instance evaluates foundation models' ability to perform context-aware Physical AI decision-making under realistic partial observability while preventing access to future information.

\subsubsection{Benchmark Instance Construction}

\textit{PhysAI-Bench} is constructed from the AI conversational episode corpus generated by the $\alpha^3$-Bench framework \cite{11609268}. Each episode represents a complete autonomous mission executed by a Physical AI agent through sequential perception, reasoning, and action while interacting with its environment using structured dialogues, MCP tool invocations, and A2A communication. Rather than evaluating complete conversational missions, \textit{PhysAI-Bench} transforms these episodes into a collection of independent decision-making problems by identifying every valid agent decision encountered during mission execution. For each selected decision point, the episode is truncated immediately before the corresponding agent action, ensuring that only the information available at decision time is exposed to the evaluated model and completely preventing future information leakage.
Formally, an AI conversational episode generated by 
$\alpha^3$-Bench is represented as
\begin{equation}
    \Gamma = (s_0, D, s_f, \delta),
\end{equation}
where $s_0$ and $s_f$ denote the initial and final Physical AI
states, respectively; $D = \{d_t\}_{t=1}^{T}$ represents the
conversational dialogue; and $\delta$ contains the episode metadata. Each dialogue turn is defined as
\begin{equation}
d_t=
\left(
r_t,\,
i_t,\,
a_t,\,
o_t,\,
n_t
\right),
\label{eq:dialog_turn}
\end{equation}
where $r_t$ denotes the speaker role, $i_t$ the high-level intent, $a_t$ the executed action(s), $o_t$ the corresponding observations, and $n_t$ the network state.

The benchmark generation process is formulated as
\begin{equation}
\Psi:
(\Gamma,\mathcal{R})
\rightarrow
Q_{\Gamma},
\label{eq:transformation}
\end{equation}
where $\mathcal{R}$ denotes the benchmark generation rules, including decision-point identification, episode truncation, context summarization, candidate action generation, rationale generation, and schema validation, and
\begin{equation}
Q_{\Gamma}=
\left\{
\tau_1,\tau_2,\ldots,\tau_M
\right\}
\label{eq:question_set}
\end{equation}
is the set of benchmark instances generated from conversational episode $\Gamma$. Since a single conversational episode may contain multiple valid agent decisions, one episode can generate multiple benchmark instances. Consequently, the complete benchmark is defined as
\begin{equation}
    \mathcal{B}
    =
    \bigcup_{\Gamma \in \mathcal{D}_{\mathrm{ep}}} Q_{\Gamma},
    \label{eq:complete_benchmark}
\end{equation}
where $\mathcal{D}_{\mathrm{ep}}$ denotes the corpus of AI
conversational episodes generated by the $\alpha^3$-Bench framework.

Each benchmark instance is represented as
\begin{equation}
    \tau_i =
    \left(
        c_i,\,
        \mathcal{O}_i,\,
        \mathcal{C}_i,\,
        a_i^{\ast},\,
        e_i,\,
        m_i
    \right),
    \label{eq:instance}
\end{equation}
where $c_i$ denotes the natural-language instruction,
$\mathcal{O}_i$ is the context summary automatically constructed
from the truncated conversational episode, and

\begin{equation}
    \mathcal{C}_i =
    \left\{
        a_{i,1}, a_{i,2}, \ldots, a_{i,K}
    \right\}
    \subseteq \mathcal{A}
\end{equation}

is the set of candidate actions. The trace-derived reference action
is denoted by $a_i^{\ast} \in \mathcal{C}_i$, $e_i$ provides the
corresponding decision rationale, and $m_i$ stores metadata including
the source and target dialogue turns, difficulty level, Physical AI
domain, physical factors, network factors, and schema version.

The reference action $a_i^{\ast}$ is obtained directly from the
original execution trace, whereas the remaining candidate actions
are automatically synthesized as plausible non-reference
alternatives. Therefore, the transformation $\Psi$ establishes a
one-to-many mapping between AI conversational episodes and benchmark
instances, converting large-scale conversational mission traces into
a standardized collection of realistic Physical AI decision-making
problems while preserving the contextual information, temporal
dependencies, and operational semantics of the original autonomous
missions.

\begin{figure}[t]
    \centering
    \includegraphics[width=0.7\columnwidth]{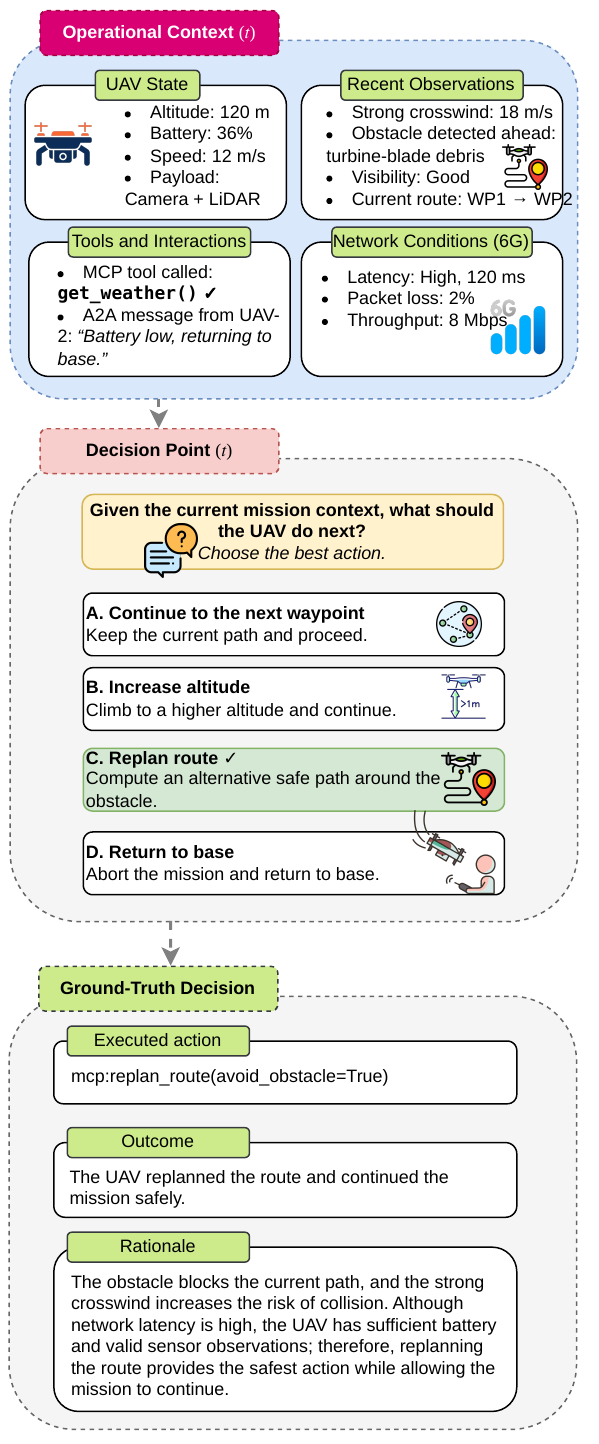}
    \caption{Structure of a \textit{PhysAI-Bench} decision instance. The model receives only the mission information available before the selected decision and chooses one of four candidate actions; its prediction is compared with the trace-derived reference action.}
    \label{fig:physai_example_instance}
\end{figure}

Fig.~\ref{fig:physai_example_instance} illustrates a representative \textit{PhysAI-Bench} instance. The evaluated model receives the operational context available immediately before the selected decision point, including the UAV state, recent observations, MCP tool interactions, A2A messages, and network conditions. It must then select the most appropriate action from four candidates. The correct action is obtained directly from the original autonomous execution trace.

Algorithm~\ref{alg:physai_generation} summarizes the implementation of the transformation $\Psi$. Starting from an AI conversational episode, the algorithm first identifies all valid agent decision points, truncates the episode immediately before each selected decision, constructs a structured context summary from the visible perception--reasoning--action history, derives the admissible action catalog from the available MCP tools, A2A interactions, and high-level actions, and then generates a multiple-choice decision problem consisting of one ground-truth action and several plausible distractors. Finally, we enrich each generated instance with a decision rationale and metadata, then validate it against the canonical \textit{PhysAI-Bench} schema. Consequently, a single AI conversational episode can generate multiple benchmark instances, establishing the one-to-many mapping defined by (\ref{eq:transformation}) while preserving the original mission's contextual information, temporal dependencies, and operational semantics.

\begin{algorithm}[t]
\caption{Generation of \textit{PhysAI-Bench} Instances}
\label{alg:physai_generation}
\begin{algorithmic}[1]

\Require AI conversational episode
$\Gamma=(s_0,D,s_f,\delta)$,
generation rules $\mathcal{R}$,
and generator set $\mathcal{G}$

\Ensure Question set
$Q_{\Gamma}=\{\tau_1,\tau_2,\ldots,\tau_M\}$

\State $Q_{\Gamma}\gets\emptyset$

\State Identify valid agent decision turns:
\Statex \hspace{\algorithmicindent}
$\mathcal{I}_{\Gamma}\gets
\{t\in\{1,\ldots,T\}\mid
r_t=\mathrm{agent}\land a_t\neq\emptyset\}$

\ForAll{$t\in\mathcal{I}_{\Gamma}$}

    \State Set target turn $t_{\mathrm{tar}}\gets t$
    \State Set source turn $t_{\mathrm{src}}\gets t-1$

    \State Truncate the dialogue:
    \Statex \hspace{\algorithmicindent}
    $D_{<t}\gets\{d_1,d_2,\ldots,d_{t-1}\}$

    \State Select an instance generator:
    \Statex \hspace{\algorithmicindent}
    $g\gets\operatorname{SelectGenerator}(\mathcal{G})$

    \State Generate the instruction and visible context:
    \Statex \hspace{\algorithmicindent}
    $(c,O)\gets
    G_{\mathrm{ctx}}^{g}(s_0,D_{<t},\mathcal{R})$

    \State Freeze the generated instruction and context:
    \Statex \hspace{\algorithmicindent}
    $\operatorname{Freeze}(c,O)$

    \State Derive the admissible action catalogue:
    \Statex \hspace{\algorithmicindent}
    $\mathcal{A}\gets
    \operatorname{DeriveActions}(s_0,D_{<t},\mathcal{R})$

    \State Generate an answer-blind candidate pool:
    \Statex \hspace{\algorithmicindent}
    $P\gets
    G_{\mathrm{cand}}^{g}(c,O,\mathcal{A},\mathcal{R})$

    \State Freeze the candidate pool:
    \Statex \hspace{\algorithmicindent}
    $\operatorname{Freeze}(P)$

    \State Extract the trace-derived reference action:
    \Statex \hspace{\algorithmicindent}
    $a^{*}\gets a_t$

    \State Validate and assemble the final candidate set:
    \Statex \hspace{\algorithmicindent}
    $C\gets
    V(O,P,a^{*},\mathcal{R})$

    \If{$C\neq\emptyset$}

        \State Randomize the position of $a^{*}$ in $C$

        \State Generate the decision rationale:
        \Statex \hspace{\algorithmicindent}
        $e\gets
        G_{\mathrm{rat}}^{g}(c,O,C,a^{*},\mathcal{R})$

        \State Construct the metadata:
        \Statex \hspace{\algorithmicindent}
        $m\gets
        \operatorname{Metadata}
        (\delta,t_{\mathrm{src}},t_{\mathrm{tar}},g,\mathcal{R})$

        \State Form the benchmark instance:
        \Statex \hspace{\algorithmicindent}
        $\tau\gets(c,O,C,a^{*},e,m)$

        \If{$\operatorname{Validate}(\tau,\mathcal{R})=\mathrm{true}$}
            \State $Q_{\Gamma}\gets Q_{\Gamma}\cup\{\tau\}$
        \EndIf

    \EndIf

\EndFor

\State \Return $Q_{\Gamma}$

\end{algorithmic}
\end{algorithm}

\subsubsection{Question Generation Pipeline}

For each instance generated by Algorithm~\ref{alg:physai_generation}, \textit{PhysAI-Bench} uses one of two advanced AI agents: DeepSeek V4.1 and Qwen3.8-Max. These agents use chain-of-thought prompting to construct standardized multiple-choice questions. The benchmark does not store or expose their internal reasoningring evaluation. None of the three generator models is included in the performance evaluation. This separation prevents direct self-evaluation and reduces generator-specific bias. The instance metadata retains the generator identity.

The pipeline separates context construction, candidate generation, and reference-action assignment. The first two stages remain blind to the action executed in the original trace. Let

\begin{equation}
\chi_i^{\mathrm{ctx}}
=
\left(s_0,D_{<t},\mathcal{R}\right)
\label{eq:context_generation_input}
\end{equation}

denote the input to the context-generation stage. Here, $s_0$ is the initial Physical AI state, $D_{<t}$ is the dialogue observed before decision step $t$, and $\mathcal{R}$ contains the benchmark-generation rules. The instruction and context summary are generated as

\begin{equation}
\left(c_i,O_i\right)
=
G_{\mathrm{ctx}}
\left(\chi_i^{\mathrm{ctx}}\right).
\label{eq:context_generation}
\end{equation}

The generator $G_{\mathrm{ctx}}$ does not receive the executed action or any subsequent trace event. The resulting instruction $c_i$ and context $O_i$ are stored and frozen before the next stage. The context contains the mission state, operational constraints, sensor observations, visible dialogue history, MCP tool responses, A2A interactions, and network conditions available at decision time.

Next, an answer-blind candidate generator constructs a pool of admissible actions:

\begin{equation}
P_i
=
G_{\mathrm{cand}}
\left(c_i,O_i,\mathcal{A}_i,\mathcal{R}\right),
\label{eq:candidate_generation}
\end{equation}

where $\mathcal{A}_i$ denotes the admissible high-level actions, MCP tools, and A2A operations. The generator $G_{\mathrm{cand}}$ does not receive the trace-derived reference action. It therefore cannot construct alternatives by directly observing the expected answer.

After the context and candidate pool are frozen, an independent validation and assembly module introduces the trace-derived reference action:

\begin{equation}
C_i
=
V\left(O_i,P_i,a_i^{*},\mathcal{R}\right).
\label{eq:candidate_validation}
\end{equation}

Here, $a_i^{*}$ denotes the action executed in the original mission trace. The validator $V(\cdot)$ constructs the final candidate set $C_i$. It confirms that $a_i^{*}$ appears exactly once and rejects semantically equivalent alternatives. It also checks that the remaining options are plausible within the visible context. The notation $a_i^{*}$ therefore represents a trace-derived reference action. It does not imply that the action is the only theoretically valid or globally optimal decision.

The decision rationale is generated only after the candidate set has been finalized:

\begin{equation}
e_i
=
G_{\mathrm{rat}}
\left(c_i,O_i,C_i,a_i^{*},\mathcal{R}\right).
\label{eq:rationale_generation}
\end{equation}

The rationale $e_i$ explains the trace-derived decision using the visible operational context. We store it for interpretation and quality assurance but do not provide it to evaluated models.

\begin{tcolorbox}[
title=\textbf{\textit{PhysAI-Bench} Benchmark Instance Format},
colback=gray!5,
colframe=black,
fonttitle=\bfseries,
boxrule=0.6pt,
arc=1mm]

\textbf{Instruction ($c_i$)}

\emph{Decision-making question}

\medskip

\textbf{Context Summary ($O_i$)}

\begin{itemize}
    \item Physical AI state
    \item Mission objectives and constraints
    \item Available MCP tools and A2A tasks
    \item Visible perception--reasoning--action trace
    \item Sensor observations and network conditions
\end{itemize}

\textbf{Action Candidates ($C_i$)}

\begin{tabular}{ll}
A. & \texttt{mcp:navigate\_to(...)}\\
B. & \texttt{mcp:hold\_position()}\\
C. & \texttt{a2a:request\_peer\_assistance(...)}\\
D. & \texttt{mcp:collision\_avoidance\_scan()}
\end{tabular}

\medskip

\textbf{Trace-Derived Reference Action ($a_i^{*}$)}

\emph{Candidate label (A--D)}

\medskip

\textbf{Decision Rationale ($e_i$)}

\emph{Explanation based only on the visible decision context}

\medskip

\textbf{Metadata ($m_i$)}

\emph{Difficulty $\bullet$ Physical Factors $\bullet$ Network Factors
$\bullet$ Source Episode $\bullet$ Generator Identity}

\end{tcolorbox}

\begin{figure*}[t]
    \centering
    \begin{subfigure}[t]{0.32\textwidth}
        \centering
        \includegraphics[width=\linewidth]
        {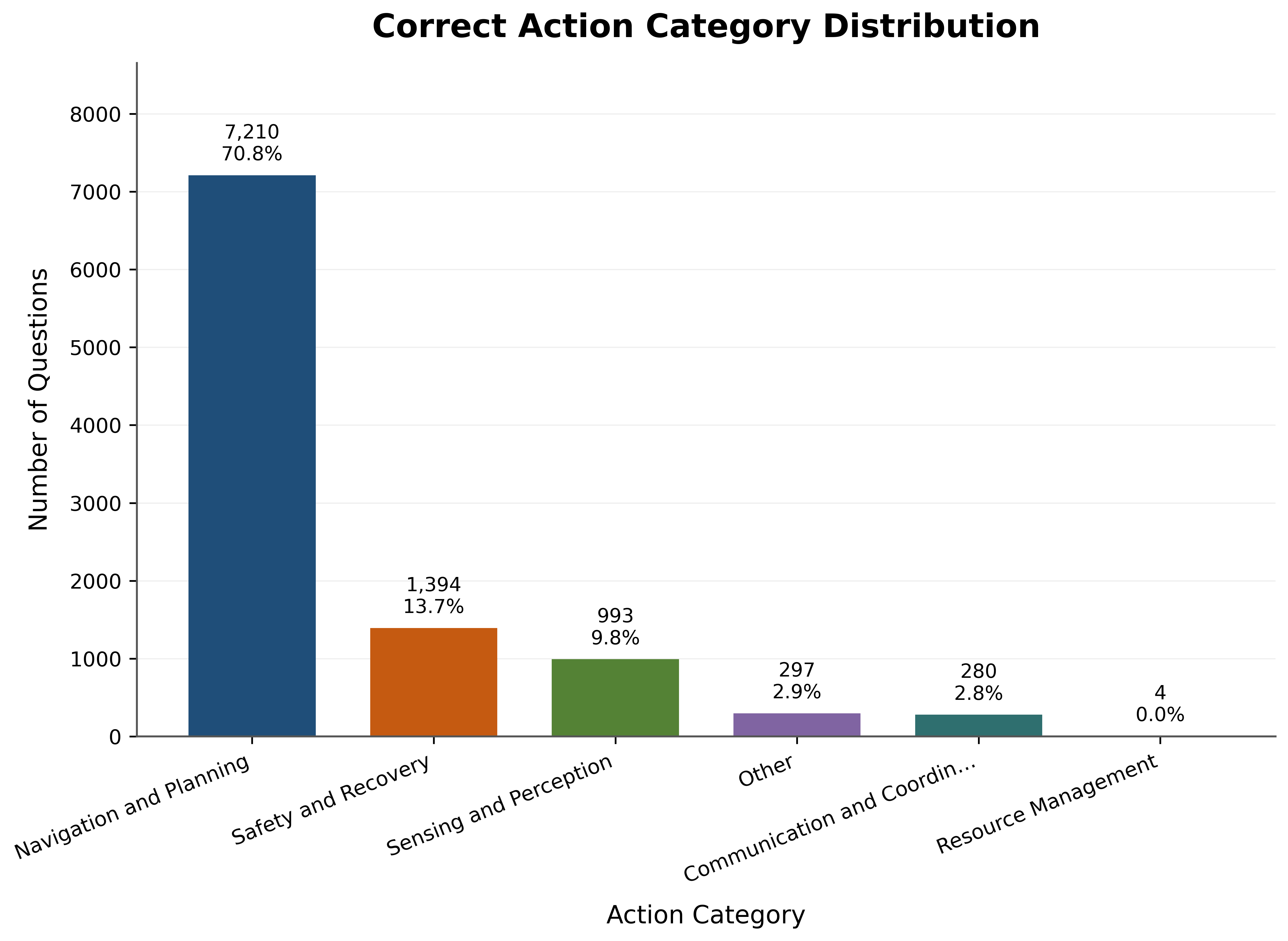}
        \caption{Correct-action categories.}
        \label{fig:action-categories}
    \end{subfigure}
    \hfill
    \begin{subfigure}[t]{0.32\textwidth}
        \centering
        \includegraphics[width=\linewidth]
        {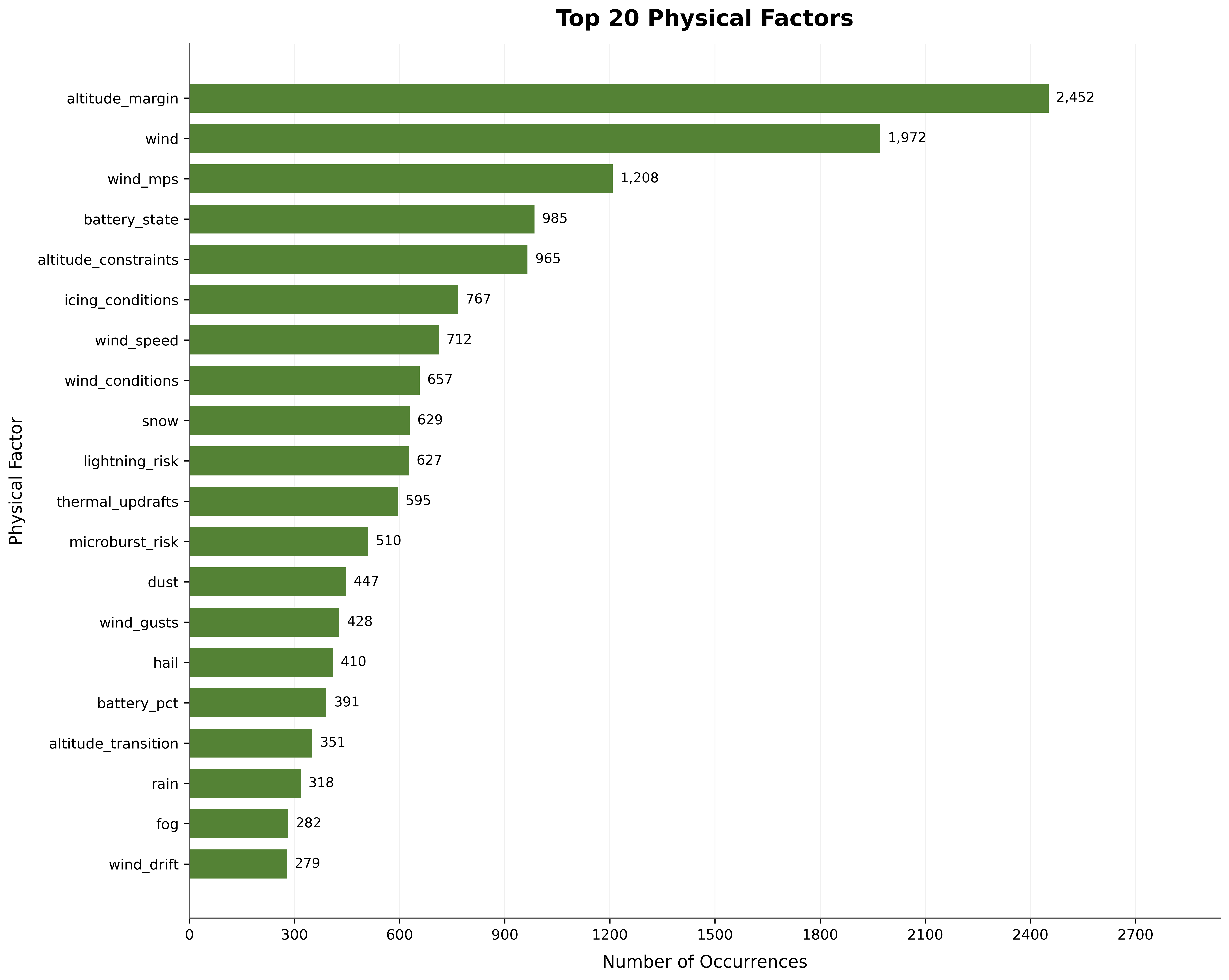}
        \caption{Most frequent physical factors.}
        \label{fig:physical-factors}
    \end{subfigure}
    \hfill
    \begin{subfigure}[t]{0.32\textwidth}
        \centering
        \includegraphics[width=\linewidth]
        {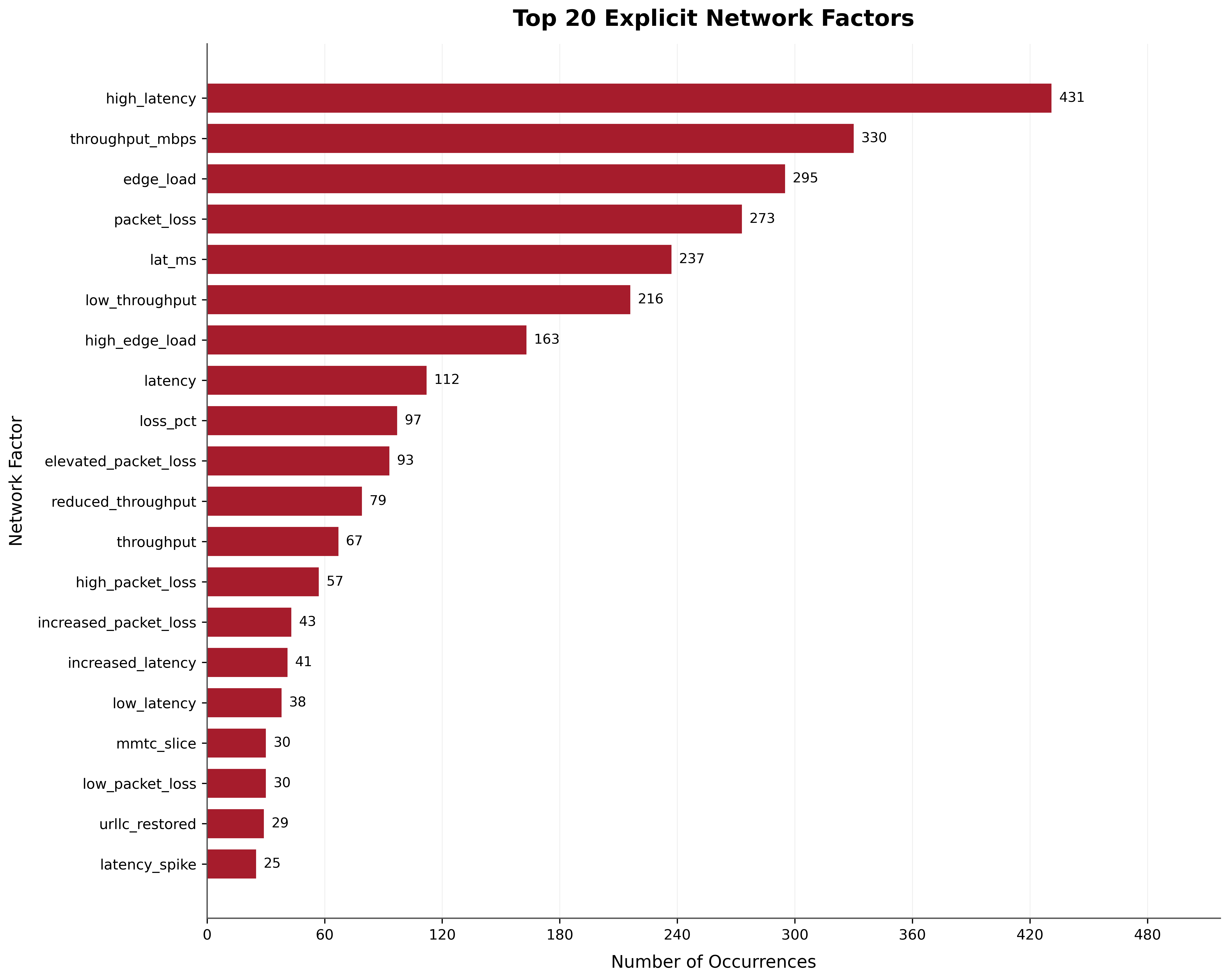}
        \caption{Most frequent network factors.}
        \label{fig:network-factors}
    \end{subfigure}

    \caption{Composition and operating-condition coverage of
    PhysAI-Bench: (a) correct-action categories, (b) physical
    factors, and (c) network factors.}
    \label{fig:benchmark-composition}
\end{figure*}

\subsection{Benchmark Composition and Coverage}

PhysAI-Bench contains 10,178 trace-derived decision-making
instances. The reference-action labels are nearly uniform across
the four candidate positions: A, B, C, and D account for 2,521
(24.8\%), 2,545 (25.0\%), 2,573 (25.3\%), and 2,539 (24.9\%)
instances, respectively, thereby limiting answer-position bias.
As shown in Fig.~\ref{fig:action-categories}, navigation and
planning constitute the largest action category, with 7,210
instances (70.8\%), followed by safety and recovery with 1,394
(13.7\%) and sensing and perception with 993 (9.8\%). The
remaining instances cover communication and coordination,
specialized decisions, and resource management. Thus, the
benchmark spans several autonomous capabilities, although its
composition is primarily oriented toward navigation- and
planning-related decisions.

Figures~\ref{fig:physical-factors} and
\ref{fig:network-factors} summarize the operating conditions
represented in the decision contexts. The most frequent physical
factors include \texttt{altitude\_margin} (2,452 occurrences),
general wind conditions (1,972), wind-speed measurements
(1,208), battery state (985), and altitude constraints (965).
The network contexts are led by high latency (431 occurrences),
throughput measurements (330), edge-computing load (295),
packet loss (273), and latency measurements (237). Together,
these distributions show that PhysAI-Bench evaluates action
selection under coupled mission, environmental, vehicle-state,
and communication constraints characteristic of UAV-centric
Physical AI. Detailed action-level, rationale, text-length,
sensor-state, and network-slice distributions can be reported
in the supplementary material or dataset documentation.

\begin{table*}[t]
\centering
\caption{Foundation Models, Characteristics, and Evaluation-Time Inference Settings.}
\label{tab:models_and_inference}
\scriptsize
\setlength{\tabcolsep}{2.5pt}
\renewcommand{\arraystretch}{1.05}
\resizebox{\textwidth}{!}{%
\begin{tabular}{lllccclcccc}
\toprule
\multicolumn{6}{c}{\textbf{Model Characteristics}} &
\multicolumn{5}{c}{\textbf{Evaluation-Time Inference}} \\
\cmidrule(lr){1-6}\cmidrule(lr){7-11}
\textbf{Model} &
\textbf{Company} &
\textbf{Category} &
\textbf{Open?} &
\textbf{Params (B)$^{\dagger}$} &
\textbf{Context} &
\textbf{Provider} &
\textbf{Input (\$/1M)} &
\textbf{Output (\$/1M)} &
\textbf{Latency (s)} &
\textbf{Throughput (tok/s)} \\
\midrule
Grok 4.5 & xAI & Frontier reasoning & Prop. & -- & 500k & xAI & 2.00 & 6.00 & 1.10 & 53 \\
Claude Fable 5 & Anthropic & Mythos reasoning & Prop. & -- & 1,000k & Claude Platform on AWS & 10.00 & 50.00 & 6.11 & 44 \\
Qwen3.7-Max & Alibaba (Qwen) & General reasoning & Prop. & -- & 1,000k & Alibaba Cloud International & 1.475 & 4.425 & 1.16 & 37 \\
NVIDIA Nemotron 3 Nano Omni 30B A3B & NVIDIA & Multimodal reasoning (MoE) & Open & 29 (3 act.) & 1,048k & NVIDIA & Free & Free & 0.62 & 88 \\
Gemma 4 26B A4B IT & Google & Multimodal reasoning & Open & 26 (4 act.) & 262k & Cloudflare & 0.10 & 0.30 & 0.44 & 51 \\
Gemma 4 31B IT & Google & Multimodal reasoning & Open & 31 & 256k & CoreWeave & 0.10 & 0.34 & 0.72 & 39 \\
GPT-5.4 Nano & OpenAI & General / small & Prop. & -- & 400k & Azure & 0.20 & 1.25 & 1.23 & 41 \\
GPT-5.3 Chat & OpenAI & General reasoning & Prop. & -- & 128k & OpenAI & 1.75 & 14.00 & 1.35 & 41 \\
Kimi K2.5 & Moonshot AI & Multimodal reasoning (MoE) & Open & 1000 (32 act.) & 262k & DeepInfra & 0.45 & 2.25 & 0.70 & 47 \\
GPT-5.2 Chat & OpenAI & General reasoning & Prop. & -- & 128k & OpenAI & 1.75 & 14.00 & 1.48 & 48 \\
Qwen3.7-Plus & Alibaba (Qwen) & General reasoning & Prop. & -- & 262k & Alibaba Cloud International & 0.32 & 1.28 & 0.91 & 12 \\
Ministral 14B 2512 & Mistral AI & General / multimodal & Open & 14 & 262k & Mistral & 0.20 & 0.20 & 0.41 & 10 \\
Ministral 8B 2512 & Mistral AI & General / multimodal & Open & 8 & 262k & NextBit & 0.30 & 0.30 & 0.65 & 4 \\
Ministral 3B 2512 & Mistral AI & General / multimodal & Open & 3 & 131k & Mistral & 0.10 & 0.10 & 0.35 & 36 \\
DeepSeek V3.2 & DeepSeek AI & Reasoning (MoE) & Open & 671 (37 act.) & 131k & StreamLake & 0.2145 & 0.3218 & 1.59 & 20 \\
Granite 4.0 H-Micro & IBM & Hybrid dense & Open & 3 & 131k & Cloudflare & 0.017 & 0.112 & 0.62 & 29 \\
Claude Haiku 4.5 & Anthropic & General reasoning & Prop. & -- & 200k & Google Vertex & 1.00 & 5.00 & 0.47 & 70 \\
Nous Hermes 4 405B & Nous Research & Hybrid reasoning & Open & 405 & 131k & Nebius Token Factory & 1.00 & 3.00 & 0.33 & 36 \\
Nous Hermes 4 70B & Nous Research & Hybrid reasoning & Open & 70 & 131k & Nebius Token Factory & 0.13 & 0.40 & 0.21 & 83 \\
Qwen3-235B & Alibaba (Qwen) & Reasoning (MoE) & Open & 235 (22 act.) & 262k & Alibaba Cloud International & 0.455 & 1.82 & 0.35 & 62 \\
Mistral Small 2603 & Mistral AI & General reasoning & Open & 24 & 128k & Venice & 0.1875 & 0.75 & 0.70 & 106 \\
Gemma 3 12B IT & Google & Multimodal reasoning & Open & 12 & 128k & DeepInfra & 0.05 & 0.15 & 0.38 & 43 \\
Gemma 3 4B IT & Google & Multimodal reasoning & Open & 4 & 128k & DeepInfra & 0.05 & 0.10 & 0.46 & 28 \\
Phi-4 & Microsoft & Reasoning & Open & 14 & 16k & DeepInfra & 0.07 & 0.14 & 0.21 & 74 \\
Llama 3.3 70B Instruct & Meta & General reasoning & Open & 70 & 131k & Groq & 0.59 & 0.79 & 0.31 & 225 \\
Amazon Nova Micro 1.0 & Amazon & General-purpose & Prop. & -- & 128k & Amazon Bedrock & 0.035 & 0.14 & 0.48 & 136 \\
Nous Hermes 3 Llama 3.1 405B & Nous Research & Reasoning/instruction & Open & 405 & 131k & DeepInfra & 1.00 & 1.00 & 0.64 & 20 \\
Qwen2.5 7B Instruct & Alibaba (Qwen) & General reasoning & Open & 7 & 131k & Together & 0.30 & 0.30 & 0.28 & 56 \\
Llama 3.2 3B Instruct & Meta & General / small & Open & 3 & 131k & Parasail & 0.05 & 0.33 & 0.19 & 246 \\
Llama 3.2 1B Instruct & Meta & Tiny general & Open & 1 & 128k & Cloudflare & 0.027 & 0.201 & 0.24 & 102 \\
\bottomrule
\end{tabular}%
}

\vspace{0.25em}
\parbox{0.98\textwidth}{\footnotesize
$^{\dagger}$Parameters are reported in billions (B). For Mixture-of-Experts (MoE) models, ``total (active)'' indicates the total number of parameters and the number activated per token during inference. Context denotes the maximum supported input length. Prices, latency, and throughput correspond to the selected provider at the time of evaluation (15 Sep. 2026).
}
\end{table*}

\section{Experimental Setup}
\label{sec:sec4}

\subsection{Evaluated Large Language Models}

We evaluate \textit{PhysAI-Bench} using 30 contemporary large language models developed by 14 industrial and research organizations, including Alibaba (Qwen), Amazon, Anthropic, DeepSeek AI, Google, IBM, Meta, Microsoft, Mistral AI, Moonshot AI, NVIDIA, Nous Research, OpenAI, and xAI. As summarized in Table~\ref{tab:models_and_inference}, the portfolio comprises proprietary and open-weight systems spanning dense and Mixture-of-Experts (MoE) architectures, multimodal models, frontier reasoning systems, and compact instruction-tuned models. Representative proprietary models include Grok~4.5, Claude Fable~5, GPT-5.3 Chat, GPT-5.2 Chat, and Qwen3.7-Max, while the open-weight portfolio includes DeepSeek V3.2, Kimi K2.5, Qwen3-235B, NVIDIA Nemotron~3 Nano Omni, and models from the Gemma, Ministral, Llama, and Nous Hermes families. Among models with disclosed parameter counts, scale ranges from 1B parameters for Llama~3.2 1B Instruct to 1{,}000B total parameters for Kimi K2.5. Other large MoE models include DeepSeek V3.2, with 671B total and 37B active parameters, and Qwen3-235B, with 235B total and 22B active parameters. Maximum supported context windows range from 16k tokens for Phi-4 to approximately 1.05 million tokens for NVIDIA Nemotron~3 Nano Omni. This heterogeneous portfolio enables a broad assessment of agentic decision-making across model scales, architectural families, context capacities, and openness regimes.

Table~\ref{tab:models_and_inference} also reports the evaluation-time inference setting selected for each model, including its provider, token prices, latency, and throughput. Representative deployments include GPT-5.3 Chat and GPT-5.2 Chat through OpenAI, Grok~4.5 through xAI, Claude Fable~5 through the Claude Platform on AWS, Qwen models through Alibaba Cloud International, Llama~3.3 70B Instruct through Groq, and DeepSeek V3.2 through StreamLake. Other open-weight models were deployed through services such as Cloudflare, CoreWeave, DeepInfra, Mistral, Nebius Token Factory, NextBit, Parasail, Together, and Venice. Across the selected deployments, input prices range from free to \$10.00 per million tokens and output prices range from free to \$50.00 per million tokens, while provider-reported latency varies from 0.19\,s to 6.11\,s and throughput ranges from 4 to 246 tokens/s. These values characterize the deployment environments at the time of evaluation and should not be interpreted as intrinsic model properties. For reproducibility, the model identifier, resolved provider, inference parameters, and response metadata are retained for every held-out request, while provider-reported statistics are recorded separately from the end-to-end latency and efficiency measurements collected during benchmark execution.

\subsection{Prompting Strategies}

{\color{black}
We consider zero-shot (0-shot), three-shot (3-shot), and five-shot
(5-shot) prompting during configuration development. In the zero-shot
setting, the model receives only the task instruction and target
question. The few-shot settings prepend three or five fixed
demonstrations. The demonstrations, prompt template, question order, and output
instructions are identical across models. Demonstration instances are
excluded from development scoring, and their complete source episodes
are excluded from the held-out evaluation. Every prompt instructs the model to return exactly one option label
from $\{A,B,C,D\}$. This controlled format supports automatic parsing
while keeping the evaluated task focused on action selection.
}

\subsection{Inference Settings}

{\color{black}
During configuration development, each model is evaluated at four
decoding temperatures,
$T \in \{0.0,0.2,0.5,0.8\}$, while all other inference parameters are
held constant whenever supported by the corresponding API. Combining
the three prompting strategies with the four temperatures produces
12 candidate configurations per model. Each development configuration is repeated three times using
deterministic item-level seeds. We record the predicted option,
end-to-end latency, generated output length, parsing status, inference
status, and provider metadata for every request.

The held-out stage uses only the model-specific configuration selected
on the development set. This frozen configuration is evaluated in
three prespecified runs on the same 500 held-out questions. No prompt,
temperature, demonstration, parser, or inference parameter is changed
after held-out predictions become available. Inference latency is measured from request submission to receipt of
the complete API response and is averaged over successful requests.  Therefore, it reflects both the computation and the deployment effects of the model,
including the provider infrastructure, the routing, the network conditions, and
the generated output length.
}

\subsection{Two-Stage Evaluation Protocol}

{\color{black}
The evaluation separates configuration selection from final
performance estimation. Let $\mathcal{B}_{\mathrm{dev}}$ denote the
35-instance human-verified development set and
$\mathcal{B}_{\mathrm{test}}$ the fixed \(500\)-instance human-verified held-out set. In the first stage, all 12 prompting--temperature configurations are
evaluated on $\mathcal{B}_{\mathrm{dev}}$. For model $M$, the frozen
configuration is selected as
\begin{equation}
c_M^{\star}
=
\operatorname*{arg\,max}_{c \in \mathcal{C}}
\mathrm{Acc}_{\mathrm{dev}}(M,c),
\label{eq:configuration_selection}
\end{equation}

where
$\mathcal{C}
=
\{0,3,5\text{-shot}\}
\times
\{0.0,0.2,0.5,0.8\}$.
When configurations have identical mean accuracy, ties are resolved
using lower run-to-run variability, lower inference latency, lower
temperature, and fewer demonstrations, in that order. Before held-out sampling, semantically identical duplicate records are
collapsed using their canonical item identifiers and content
fingerprints, recovering the 10,178 unique benchmark instances.
Records sharing an identifier but containing conflicting questions or
answers are rejected rather than silently deduplicated.

The held-out candidates exclude every source episode represented in
the development set or demonstration pool. From the remaining
episodes, an episode-aware deterministic procedure selects exactly
500 questions using a fixed seed. The procedure distributes selections
across episodes to prevent long missions from dominating the sample. The same held-out item identifiers apply to every model. The
selected identifiers, source episodes, sampling seed, dataset
fingerprint, and frozen configuration are stored in a locked manifest.
Consequently, the held-out set cannot be modified in response to model
performance.
}

\begin{algorithm}[t]
\color{black}
\caption{\textit{PhysAI-Bench} Two-Stage LLM Evaluation Protocol}
\label{alg:physai_evaluation}
\begin{algorithmic}[1]
\Require Full benchmark $\mathcal{B}$, development set
$\mathcal{B}_{\mathrm{dev}}$, demonstration set
$\mathcal{B}_{\mathrm{demo}}$, model set $\mathcal{M}$,
configuration set $\mathcal{C}$, runs $R$, seed $s$
\Ensure Frozen configurations and held-out evaluation metrics

\State Remove semantically identical duplicate records from
$\mathcal{B}$
\State Exclude episodes occurring in
$\mathcal{B}_{\mathrm{dev}} \cup \mathcal{B}_{\mathrm{demo}}$
\State Construct fixed test set
$\mathcal{B}_{\mathrm{test}}$ of 500 instances using seed $s$
\State Lock the test identifiers, episode identifiers, and fingerprint

\ForAll{$M \in \mathcal{M}$}
    \ForAll{$c \in \mathcal{C}$}
        \For{$r=1$ to $R$}
            \State Evaluate $M$ on $\mathcal{B}_{\mathrm{dev}}$
            using configuration $c$
        \EndFor
        \State Aggregate development metrics across runs
    \EndFor

    \State Select $c_M^{\star}$ using
    Eq.~\eqref{eq:configuration_selection}
    \State Freeze the prompt, demonstrations, parser, and inference settings

    \For{$r=1$ to $R$}
        \ForAll{$\tau_i \in \mathcal{B}_{\mathrm{test}}$}
            \State Generate the prompt for $\tau_i$ using $c_M^{\star}$
            \State Obtain and parse the model prediction
            \State Record correctness, latency, output length,
            parsing status, and inference status
        \EndFor
    \EndFor

    \State Aggregate held-out metrics using source-episode clusters
\EndFor

\State \Return frozen configurations and held-out results
\end{algorithmic}
\end{algorithm}

\begin{figure*}[t]
\centering

\begin{minipage}[t]{0.49\textwidth}
\vspace{0pt}
\centering

\resizebox{\linewidth}{!}{%
\begin{tikzpicture}
\begin{axis}[
  leaderboardaxis,
  yticklabels={
    Gemma 3 12B IT,
    Claude Fable 5,
    MoonshotAI Kimi K2.5,
    Nous Hermes 4 70B,
    GPT-5.4 Nano,
    NVIDIA Nemotron 3 Omni 30B,
    Ministral 8B 2512,
    Llama 3.2 3B Instruct,
    Amazon Nova Micro V1,
    Microsoft Phi-4,
    Gemma 3 4B IT,
    IBM Granite 4.0 H Micro,
    Qwen 2.5 7B Instruct,
    Llama 3.2 1B Instruct,
    Ministral 3B 2512
  }
]

\addplot[leaderboardbar]
table[
  row sep=\\,
  x=mean,
  y=idx,
  x error=sd
] {
idx mean sd\\
0 55.56 3.50\\
1 55.24 1.65\\
2 54.29 4.95\\
3 53.54 1.75\\
4 52.53 9.26\\
5 46.46 3.50\\
6 45.45 0.00\\
7 44.44 1.75\\
8 42.86 0.00\\
9 41.90 4.36\\
10 41.41 1.75\\
11 40.95 1.65\\
12 38.10 1.65\\
13 32.32 14.00\\
14 30.30 3.03\\
};

\leadercfg{0}{3s, $T=.5$}
\leadercfg{1}{3s, $T=.2$}
\leadercfg{2}{3s, $T=.2$}
\leadercfg{3}{5s, $T=.2$}
\leadercfg{4}{5s, $T=.0$}
\leadercfg{5}{5s, $T=.5$}
\leadercfg{6}{5s, $T=.0$}
\leadercfg{7}{5s, $T=.0$}
\leadercfg{8}{5s, $T=.0$}
\leadercfg{9}{0s, $T=.5$}
\leadercfg{10}{5s, $T=.5$}
\leadercfg{11}{5s, $T=.2$}
\leadercfg{12}{3s, $T=.8$}
\leadercfg{13}{3s, $T=.8$}
\leadercfg{14}{0s, $T=.0$}

\leadermean{0}{55.56}{3.50}{55.56}
\leadermean{1}{55.24}{1.65}{55.24}
\leadermean{2}{54.29}{4.95}{54.29}
\leadermean{3}{53.54}{1.75}{53.54}
\leadermean{4}{52.53}{9.26}{52.53}
\leadermean{5}{46.46}{3.50}{46.46}
\leadermean{6}{45.45}{0.00}{45.45}
\leadermean{7}{44.44}{1.75}{44.44}
\leadermean{8}{42.86}{0.00}{42.86}
\leadermean{9}{41.90}{4.36}{41.90}
\leadermean{10}{41.41}{1.75}{41.41}
\leadermean{11}{40.95}{1.65}{40.95}
\leadermean{12}{38.10}{1.65}{38.10}
\leadermean{13}{32.32}{14.00}{32.32}
\leadermean{14}{30.30}{3.03}{30.30}

\end{axis}
\end{tikzpicture}%
}

\par\smallskip
{\small (b) Models ranked 16--30.}

\end{minipage}
\hfill
%
\begin{minipage}[t]{0.49\textwidth}
\vspace{0pt}
\centering

\resizebox{\linewidth}{!}{%
\begin{tikzpicture}
\begin{axis}[
  leaderboardaxis,
  yticklabels={
    \textbf{xAI Grok 4.5},
    \textbf{DeepSeek V3.2},
    \underline{Claude Haiku 4.5},
    Qwen3.7-Plus,
    GPT-5.2 Chat,
    Nous Hermes 4 405B,
    Nous Hermes 3 Llama 3.1 405B,
    Mistral Small 2603,
    Gemma 4 31B IT,
    GPT-5.3 Chat,
    Qwen3.7-Max,
    Gemma 4 26B A4B IT,
    Qwen3-235B,
    Ministral 14B 2512,
    Llama 3.3 70B Instruct
  }
]

\addplot[leaderboardbar]
table[
  row sep=\\,
  x=mean,
  y=idx,
  x error=sd
] {
idx mean sd\\
0 70.48 3.30\\
1 70.48 1.65\\
2 68.57 0.00\\
3 67.68 3.50\\
4 67.68 1.75\\
5 67.68 1.75\\
6 67.68 1.75\\
7 67.09 1.71\\
8 66.67 0.00\\
9 66.67 3.03\\
10 65.66 1.75\\
11 64.65 1.75\\
12 63.64 0.00\\
13 60.61 3.03\\
14 56.57 1.75\\
};
\leadercfg{0}{3s, $T=.5$}
\leadercfg{1}{3s, $T=.0$}
\leadercfg{2}{5s, $T=.0$}
\leadercfg{3}{3s, $T=.8$}
\leadercfg{4}{5s, $T=.0$}
\leadercfg{5}{5s, $T=.0$}
\leadercfg{6}{5s, $T=.2$}
\leadercfg{7}{5s, $T=.0$}
\leadercfg{8}{5s, $T=.5$}
\leadercfg{9}{3s, $T=.2$}
\leadercfg{10}{3s, $T=.2$}
\leadercfg{11}{3s, $T=.2$}
\leadercfg{12}{3s, $T=.8$}
\leadercfg{13}{5s, $T=.8$}
\leadercfg{14}{5s, $T=.8$}

\leadermean{0}{70.48}{3.30}{\textbf{70.48}}
\leadermean{1}{70.48}{1.65}{\textbf{70.48}}
\leadermean{2}{68.57}{0.00}{\underline{68.57}}
\leadermean{3}{67.68}{3.50}{67.68}
\leadermean{4}{67.68}{1.75}{67.68}
\leadermean{5}{67.68}{1.75}{67.68}
\leadermean{6}{67.68}{1.75}{67.68}
\leadermean{7}{67.09}{1.71}{67.09}
\leadermean{8}{66.67}{0.00}{66.67}
\leadermean{9}{66.67}{3.03}{66.67}
\leadermean{10}{65.66}{1.75}{65.66}
\leadermean{11}{64.65}{1.75}{64.65}
\leadermean{12}{63.64}{0.00}{63.64}
\leadermean{13}{60.61}{3.03}{60.61}
\leadermean{14}{56.57}{1.75}{56.57}

\end{axis}
\end{tikzpicture}%
}

\par\smallskip
{\small (a) Models ranked 1--15.}

\end{minipage}

\caption{
Configuration selection on the 35-instance human-verified development
set. Numbers beside the bars show mean accuracy over three independent
runs, while whiskers show $\pm$ one standard deviation. The notation
``3s, $T=.5$'', for example, denotes 3-shot prompting with temperature
$0.5$. \textbf{Bold} indicates the best-performing models, while
\underline{underlining} indicates the second-best model.
}
\label{fig:leaderboard}

\end{figure*}

\subsection{Evaluation Metrics}

{\color{black}
The development stage serves two purposes: selecting a configuration
for each model and measuring sensitivity to prompting and temperature.
The existing 35-instance results are therefore reported as
development-set analyses rather than as estimates of held-out
generalization. For every development configuration, we report the mean decision
accuracy and sample standard deviation across three runs. The
corresponding Student-$t$ interval describes uncertainty in the
three-run mean on the fixed development questions; it does not
represent uncertainty across the full benchmark population.

The primary performance estimate is strict decision accuracy on
$\mathcal{B}_{\mathrm{test}}$. A prediction is correct only when the
parsed option matches the reference label. Parsing and inference
failures count as incorrect under this metric and are additionally
reported through separate parse-success and inference-success rates. Because multiple questions may originate from the same autonomous
mission, individual test instances are not treated as statistically
independent. Held-out 95\% confidence intervals are computed through
cluster bootstrap resampling at the source-episode level. All
questions and prespecified inference runs associated with a sampled
episode remain within the same bootstrap cluster.

To quantify configuration-selection bias, we define the optimism gap
for model $M$ as
\begin{equation}
\Delta_M
=
\mathrm{Acc}_{\mathrm{dev}}(M,c_M^{\star})
-
\mathrm{Acc}_{\mathrm{test}}(M,c_M^{\star}).
\label{eq:optimism_gap}
\end{equation}
A positive $\Delta_M$ indicates that the best development
configuration achieved higher accuracy on the selection set than on
previously unseen decision points. Secondary metrics include run-to-run variability, worst-case
development accuracy, configuration range, inference latency, output
length, parse-success rate, inference-success rate, and
Accuracy/Second. These measurements characterize configuration
sensitivity, computational efficiency, and output reliability, while
the fixed held-out accuracy remains the primary basis for comparing
models.
}

\section{Experimental Results}
\label{sec:experimental_Results}

\subsection{Model-Specific Configuration Selection}

Figure~\ref{fig:leaderboard} summarizes the best performance achieved by each evaluated LLM under its optimal prompting strategy and decoding temperature. Overall, \textit{PhysAI-Bench} remains a challenging benchmark, with the highest accuracy reaching only 70.48\%, indicating that current frontier LLMs are still far from reliably solving agentic decision-making tasks. xAI Grok 4.5 and DeepSeek V3.2 jointly achieved the highest accuracy of 70.48\%. Although both models reached the same peak performance, DeepSeek V3.2 exhibited substantially lower variability ($\pm$1.65) than xAI Grok 4.5 ($\pm$3.30), indicating more stable behaviour across repeated runs. Claude Haiku 4.5 achieved the next-highest accuracy (68.57$\pm$0.00\%), with perfectly consistent performance across all runs. This was followed by four models, namely Qwen3.7-Plus, GPT-5.2 Chat, Nous Hermes 4 405B, and Nous Hermes 3 Llama 3.1 405B, each achieving 67.68\%. Among these, Qwen3.7-Plus exhibited a larger standard deviation ($\pm$3.50) than the other three models ($\pm$1.75), suggesting lower stability despite identical peak accuracy. Other strong performers include Mistral Small 2603 (67.09$\pm$1.71\%), Gemma 4 31B IT (66.67$\pm$0.00\%), GPT-5.3 Chat (66.67$\pm$3.03\%), Qwen3.7-Max (65.66\%), Gemma 4 26B A4B IT (64.65\%), and Qwen3-235B (63.64\%), while Ministral 14B 2512 remained the last model exceeding the 60\% accuracy threshold.

A noticeable performance degradation is observed for smaller and earlier-generation models. NVIDIA Nemotron 3 Nano Omni 30B A3B, Ministral 8B 2512, Llama 3.2 3B Instruct, Amazon Nova Micro V1, Gemma 3 4B IT, IBM Granite 4.0 H Micro, Llama 3.2 1B Instruct, and Ministral 3B 2512 obtained only 30.30--46.46\% accuracy. In particular, Llama 3.2 1B Instruct exhibited the highest variability ($\pm$14.00), with a confidence interval spanning [0.00, 67.09], indicating unstable behavior across repeated runs. These results suggest that increasing parameter count alone does not guarantee strong performance on \textit{PhysAI-Bench}. Instead, models specifically optimized for reasoning, instruction following, and agent-oriented behaviour consistently outperform considerably larger models that are not explicitly tuned for such capabilities.

\begin{figure}[t]
\centering

{\scriptsize
\tikz[baseline=-0.5ex]
  \draw[promptblue,line width=0.7pt]
  (0,0) circle (1.5pt);
\;0-shot
\qquad
\tikz[baseline=-0.5ex]
  \fill[promptblue]
  (-1.4pt,-1.4pt) rectangle (1.4pt,1.4pt);
\;3-shot
\qquad
\tikz[baseline=-0.5ex]
  \fill[promptblue]
  (0,1.8pt) -- (-1.7pt,-1.4pt) --
  (1.7pt,-1.4pt) -- cycle;
\;5-shot
}
\par\vspace{1mm}

%
\begin{minipage}[t]{0.4\textwidth}
\vspace{0pt}
\centering

\resizebox{\linewidth}{!}{%
\begin{tikzpicture}
\begin{axis}[
  promptaxis,
  ymin=-0.7,
  ymax=14.55,
  ytick={0,...,14},
  yticklabels={
    xAI Grok 4.5,
    DeepSeek V3.2,
    Claude Haiku 4.5,
    GPT-5.2 Chat,
    Nous Hermes 4 405B,
    Nous Hermes 3 Llama 3.1 405B,
    Mistral Small 2603,
    Gemma 4 31B IT,
    GPT-5.3 Chat,
    Qwen3.7-Max,
    Gemma 4 26B A4B IT,
    Qwen3-235B,
    Ministral 14B 2512,
    Llama 3.3 70B Instruct,
    Gemma 3 12B IT
  }
]

\promptconnect{0}{57.14}{69.52}{65.00}
\promptconnect{1}{43.81}{68.57}{61.67}
\promptconnect{2}{51.19}{63.33}{68.33}
\promptconnect{3}{43.94}{61.87}{64.39}
\promptconnect{4}{41.67}{60.61}{66.42}
\promptconnect{5}{54.29}{61.62}{66.67}
\promptconnect{6}{40.00}{55.72}{64.44}
\promptconnect{7}{53.28}{61.11}{65.15}
\promptconnect{8}{55.56}{64.90}{63.14}
\promptconnect{9}{55.30}{60.10}{59.85}
\promptconnect{10}{55.56}{63.64}{58.33}
\promptconnect{11}{48.48}{61.36}{59.84}
\promptconnect{12}{34.12}{50.25}{57.58}
\promptconnect{13}{30.05}{32.58}{55.81}
\promptconnect{14}{34.09}{53.71}{51.01}

\addplot[
  only marks,
  mark=o,
  mark size=2.1pt,
  color=promptblue,
  line width=0.7pt
]
coordinates {
  (57.14,0)
  (43.81,1)
  (51.19,2)
  (43.94,3)
  (41.67,4)
  (54.29,5)
  (40.00,6)
  (53.28,7)
  (55.56,8)
  (55.30,9)
  (55.56,10)
  (48.48,11)
  (34.12,12)
  (30.05,13)
  (34.09,14)
};

\addplot[
  only marks,
  mark=square*,
  mark size=1.9pt,
  color=promptblue
]
coordinates {
  (69.52,0)
  (68.57,1)
  (63.33,2)
  (61.87,3)
  (60.61,4)
  (61.62,5)
  (55.72,6)
  (61.11,7)
  (64.90,8)
  (60.10,9)
  (63.64,10)
  (61.36,11)
  (50.25,12)
  (32.58,13)
  (53.71,14)
};

\addplot[
  only marks,
  mark=triangle*,
  mark size=2.2pt,
  color=promptblue
]
coordinates {
  (65.00,0)
  (61.67,1)
  (68.33,2)
  (64.39,3)
  (66.42,4)
  (66.67,5)
  (64.44,6)
  (65.15,7)
  (63.14,8)
  (59.85,9)
  (58.33,10)
  (59.84,11)
  (57.58,12)
  (55.81,13)
  (51.01,14)
};

\draw[
  black!20,
  line width=0.35pt
]
(axis cs:75,-0.6) --
(axis cs:75,14.5);

\node[
  anchor=west,
  font=\tiny\bfseries
] at (axis cs:78,-0.43) {Gain};

\promptgain{0}{+21.67\%}
\promptgain{1}{+56.52\%}
\promptgain{2}{+33.49\%}
\promptgain{3}{+46.54\%}
\promptgain{4}{+59.41\%}
\promptgain{5}{+22.80\%}
\promptgain{6}{+61.10\%}
\promptgain{7}{+22.28\%}
\promptgain{8}{+16.81\%}
\promptgain{9}{+8.68\%}
\promptgain{10}{+14.55\%}
\promptgain{11}{+26.60\%}
\promptgain{12}{+68.75\%}
\promptgain{13}{+85.73\%}
\promptgain{14}{+57.55\%}

\end{axis}
\end{tikzpicture}%
}

\par\smallskip
{\small (a) Top 15 models.}

\end{minipage}

\begin{minipage}[t]{0.4\textwidth}
\vspace{0pt}
\centering

\resizebox{\linewidth}{!}{%
\begin{tikzpicture}
\begin{axis}[
  promptaxis,
  ymin=-0.7,
  ymax=13.55,
  ytick={0,...,13},
  yticklabels={
    Claude Fable 5,
    MoonshotAI Kimi K2.5,
    Nous Hermes 4 70B,
    GPT-5.4 Nano,
    NVIDIA Nemotron 3 Omni 30B,
    Ministral 8B 2512,
    Llama 3.2 3B Instruct,
    Amazon Nova Micro V1,
    Microsoft Phi-4,
    Gemma 3 4B IT,
    IBM Granite 4.0 H Micro,
    Qwen 2.5 7B Instruct,
    Llama 3.2 1B Instruct,
    Ministral 3B 2512
  }
]

\promptconnect{0}{49.76}{55.00}{45.95}
\promptconnect{1}{18.10}{37.62}{48.10}
\promptconnect{2}{31.08}{47.02}{49.75}
\promptconnect{3}{40.15}{50.25}{50.00}
\promptconnect{4}{44.44}{42.93}{44.44}
\promptconnect{5}{29.55}{34.09}{44.95}
\promptconnect{6}{27.27}{25.00}{40.91}
\promptconnect{7}{21.19}{36.43}{41.43}
\promptconnect{8}{37.38}{35.95}{38.33}
\promptconnect{9}{23.74}{28.03}{39.90}
\promptconnect{10}{27.86}{31.19}{38.33}
\promptconnect{11}{12.62}{37.38}{33.81}
\promptconnect{12}{27.78}{29.30}{26.26}
\promptconnect{13}{27.53}{22.73}{25.25}

\addplot[
  only marks,
  mark=o,
  mark size=2.1pt,
  color=promptblue,
  line width=0.7pt
]
coordinates {
  (49.76,0)
  (18.10,1)
  (31.08,2)
  (40.15,3)
  (44.44,4)
  (29.55,5)
  (27.27,6)
  (21.19,7)
  (37.38,8)
  (23.74,9)
  (27.86,10)
  (12.62,11)
  (27.78,12)
  (27.53,13)
};

\addplot[
  only marks,
  mark=square*,
  mark size=1.9pt,
  color=promptblue
]
coordinates {
  (55.00,0)
  (37.62,1)
  (47.02,2)
  (50.25,3)
  (42.93,4)
  (34.09,5)
  (25.00,6)
  (36.43,7)
  (35.95,8)
  (28.03,9)
  (31.19,10)
  (37.38,11)
  (29.30,12)
  (22.73,13)
};

\addplot[
  only marks,
  mark=triangle*,
  mark size=2.2pt,
  color=promptblue
]
coordinates {
  (45.95,0)
  (48.10,1)
  (49.75,2)
  (50.00,3)
  (44.44,4)
  (44.95,5)
  (40.91,6)
  (41.43,7)
  (38.33,8)
  (39.90,9)
  (38.33,10)
  (33.81,11)
  (26.26,12)
  (25.25,13)
};

\draw[
  black!20,
  line width=0.35pt
]
(axis cs:75,-0.6) --
(axis cs:75,13.5);

\node[
  anchor=west,
  font=\tiny\bfseries
] at (axis cs:78,-0.43) {Gain};

\promptgain{0}{+10.53\%}
\promptgain{1}{+165.75\%}
\promptgain{2}{+60.05\%}
\promptgain{3}{+25.17\%}
\promptgain{4}{+0.00\%}
\promptgain{5}{+52.11\%}
\promptgain{6}{+50.00\%}
\promptgain{7}{+95.52\%}
\promptgain{8}{+2.55\%}
\promptgain{9}{+68.07\%}
\promptgain{10}{+37.60\%}
\promptgain{11}{+196.20\%}
\promptgain{12}{+5.48\%}
\promptgain{13}{$-8.28\%$}

\end{axis}
\end{tikzpicture}%
}

\par\smallskip
{\small (b) Remaining 14 models.}

\end{minipage}

\caption{
Effect of prompting strategy during development-stage configuration
selection.
}
\label{fig:prompting-effect}

\end{figure}

\subsection{\textcolor{black}{Impact of Prompting Strategy}}

Figure~\ref{fig:prompting-effect} evaluates the influence of prompting strategies by averaging performance across all decoding temperatures. Overall, increasing the number of in-context examples generally improves performance, with 5-shot prompting producing the best results for most models. The highest average 5-shot accuracy was achieved by Claude Haiku 4.5 (68.33\%), followed by Nous Hermes 3 Llama 3.1 405B (66.67\%), Nous Hermes 4 405B (66.42\%), Gemma 4 31B IT (65.15\%), Mistral Small 2603 (64.44\%), and GPT-5.2 Chat (64.39\%). In contrast, xAI Grok 4.5 achieved the best 3-shot performance (69.52\%), followed by DeepSeek V3.2 (68.57\%), demonstrating that these models benefit substantially from a moderate number of in-context demonstrations. Regarding zero-shot performance, GPT-5.3 Chat and Gemma 4 26B A4B IT jointly achieved the highest average accuracy (55.56\%), closely followed by Qwen3.7-Max (55.30\%) and Nous Hermes 3 Llama 3.1 405B (54.29\%), indicating strong intrinsic reasoning capabilities without in-context examples. The largest relative improvement from zero-shot to the best few-shot configuration was observed for Qwen2.5 7B Instruct (+196.20\%), followed by Kimi K2.5 (+165.75\%) and Amazon Nova Micro V1 (+95.52\%), suggesting that these models benefit substantially from in-context learning. In contrast, NVIDIA Nemotron 3 Nano Omni 30B A3B (+0.00\%), Llama 3.2 1B Instruct (+5.48\%), and Qwen3.7-Max (+8.68\%) showed only marginal improvements. Overall, these results show that while few-shot prompting generally improves performance across modern LLMs, its effectiveness remains highly model-dependent, with different architectures showing markedly different sensitivities to increasing prompt complexity.

\begin{figure}[t]
\centering
%
\begin{minipage}[t]{0.3\textwidth}
\vspace{0pt}
\centering

\resizebox{\linewidth}{!}{%
\begin{tikzpicture}[x=1cm,y=1cm]

\path[use as bounding box]
  (-3.35,0.85) rectangle (4.85,-6.85);

\temperatureheader

\temperaturerow{0}
  {xAI Grok 4.5}
  {63.49}{64.44}{64.44}{63.17}{0.65}

\temperaturerow{1}
  {DeepSeek V3.2}
  {59.68}{58.41}{57.78}{56.19}{1.48}

\temperaturerow{2}
  {Claude Haiku 4.5}
  {60.95}{60.95}{61.27}{60.63}{0.26}

\temperaturerow{3}
  {GPT-5.2 Chat}
  {58.25}{54.88}{56.23}{57.58}{1.49}

\temperaturerow{4}
  {Nous Hermes 4 405B}
  {56.23}{56.57}{56.90}{55.22}{0.74}

\temperaturerow{5}
  {Nous Hermes 3 Llama 3.1 405B}
  {61.62}{61.28}{61.28}{59.26}{1.13}

\temperaturerow{6}
  {Mistral Small 2603}
  {54.80}{54.46}{52.54}{51.75}{1.46}

\temperaturerow{7}
  {Gemma 4 31B IT}
  {59.93}{60.27}{59.93}{59.93}{0.17}

\temperaturerow{8}
  {GPT-5.3 Chat}
  {62.97}{61.28}{60.61}{59.94}{1.30}

\temperaturerow{9}
  {Qwen3.7-Max}
  {56.90}{62.29}{57.58}{56.90}{2.60}

\temperaturerow{10}
  {Gemma 4 26B A4B IT}
  {59.60}{59.26}{58.92}{58.92}{0.33}

\temperaturerow{11}
  {Qwen3-235B}
  {57.24}{56.57}{55.22}{55.89}{0.87}

\temperaturerow{12}
  {Ministral 14B 2512}
  {46.46}{46.14}{46.46}{49.16}{1.31}

\temperaturerow{13}
  {Llama 3.3 70B Instruct}
  {39.73}{39.06}{39.06}{40.07}{0.48}

\temperaturerow{14}
  {Gemma 3 12B IT}
  {38.38}{38.38}{39.39}{37.37}{0.82}

\end{tikzpicture}%
}

\par\smallskip
{\small (a) Top 15 models.}

\end{minipage}

\begin{minipage}[t]{0.3\textwidth}
\vspace{0pt}
\centering

\resizebox{\linewidth}{!}{%
\begin{tikzpicture}[x=1cm,y=1cm]

\path[use as bounding box]
  (-3.35,0.85) rectangle (4.85,-6.85);

\temperatureheader

\temperaturerow{0}
  {Claude Fable 5}
  {50.16}{50.16}{50.48}{50.16}{0.16}

\temperaturerow{1}
  {MoonshotAI Kimi K2.5}
  {37.46}{38.10}{40.95}{21.90}{8.43}

\temperaturerow{2}
  {Nous Hermes 4 70B}
  {41.41}{42.42}{40.74}{39.04}{1.46}

\temperaturerow{3}
  {GPT-5.4 Nano}
  {48.15}{43.10}{48.15}{47.81}{2.36}

\temperaturerow{4}
  {NVIDIA Nemotron 3 Omni 30B A3B}
  {44.44}{42.09}{44.78}{42.09}{1.37}

\temperaturerow{5}
  {Ministral 8B 2512}
  {36.03}{36.03}{37.04}{35.69}{0.64}

\temperaturerow{6}
  {Llama 3.2 3B Instruct}
  {35.02}{31.65}{31.65}{27.27}{3.18}

\temperaturerow{7}
  {Amazon Nova Micro V1}
  {32.06}{33.65}{32.70}{33.65}{0.66}

\temperaturerow{8}
  {Microsoft Phi-4}
  {34.92}{35.87}{38.10}{40.00}{2.29}

\temperaturerow{9}
  {Gemma 3 4B IT}
  {29.63}{30.64}{31.65}{30.30}{0.74}

\temperaturerow{10}
  {IBM Granite 4.0 H Micro}
  {32.70}{33.65}{32.06}{31.43}{0.84}

\temperaturerow{11}
  {Qwen 2.5 7B Instruct}
  {27.94}{26.67}{27.62}{29.52}{1.06}

\temperaturerow{12}
  {Llama 3.2 1B Instruct}
  {29.29}{28.62}{25.59}{27.27}{1.60}

\temperaturerow{13}
  {Ministral 3B 2512}
  {26.26}{24.58}{24.24}{25.59}{0.90}

\end{tikzpicture}%
}

\par\smallskip
{\small (b) Remaining 14 models.}

\end{minipage}
\hfill
\caption{
Effect of decoding temperature during development-stage configuration
selection. Cell values report accuracy averaged across prompting
strategies.
}
\label{fig:temperature-effect}

\end{figure}

\subsection{\textcolor{black}{Impact of Decoding Temperature}}

Figure~\ref{fig:temperature-effect} evaluates the impact of decoding temperature by averaging model performance across all prompting strategies. Overall, the results indicate that decoding temperature has limited influence on \textit{PhysAI-Bench} performance, as most models show relatively small accuracy variations across the evaluated settings. xAI Grok 4.5 achieved the highest average accuracy at $T=0.2$ (64.44\%), $T=0.5$ (64.44\%), and $T=0.8$ (63.17\%), while GPT-5.3 Chat obtained the best performance at $T=0.0$ with 62.97\%. Nous Hermes 3 Llama 3.1 405B also maintained consistently high performance, exceeding 61\% accuracy at three temperature settings, whereas Claude Haiku 4.5 consistently ranked among the top-performing models across all decoding temperatures, achieving the second-highest performance at both $T=0.5$ (61.27\%) and $T=0.8$ (60.63\%). Regarding robustness, Claude Fable 5 exhibited the lowest standard deviation (0.16), indicating the most stable performance across decoding temperatures, closely followed by Gemma 4 31B IT (0.17). Gemma 4 26B A4B IT also demonstrated excellent stability with a standard deviation of only 0.33. In contrast, models such as Llama 3.2 3B Instruct (3.18), Qwen3.7-Max (2.60), and GPT-5.4 Nano (2.36) showed larger performance variations. These findings suggest that decoding temperature has a relatively minor impact on the strongest reasoning models, while model architecture and intrinsic reasoning capability play a much more significant role in determining overall performance on \textit{PhysAI-Bench}.

\begin{table}[t]
\centering
\caption{Development-Set Accuracy and Sensitivity Across Prompting and Decoding Configurations.}
\label{tab:agentic}
\scriptsize
\begin{tabular}{lcccc}
\toprule
Model & OA & Worst & Range & SD\\
\midrule
\textbf{xAI Grok 4.5}
&\textbf{63.89}
&\textbf{56.19}
&14.29
&2.17\\

Claude Fable 5
&50.24
&44.76
&10.48
&2.03\\

DeepSeek V3.2
&58.02
&42.86
&27.62
&4.18\\

Qwen3-235B
&56.73
&46.46
&17.17
&2.14\\

Qwen3.7-Max
&58.42
&53.54
&12.12
&3.08\\

GPT-5.2 Chat
&56.73
&42.42
&25.25
&3.76\\

\underline{GPT-5.3 Chat}
&\underline{61.20}
&53.54
&13.13
&3.00\\

GPT-5.4 Nano
&46.80
&34.34
&18.18
&4.57\\

\underline{Claude Haiku 4.5}
&60.95
&50.48
&17.14
&\underline{0.89}\\

MoonshotAI Kimi K2.5
&35.60
&0.00
&54.29
&10.97\\

Mistral Small 2603
&53.39
&36.75
&30.34
&2.27\\

Qwen 2.5 7B Instruct
&27.94
&11.43
&26.67
&0.74\\

Ministral 14B 2512
&47.36
&33.33
&27.27
&2.19\\

\underline{NVIDIA Nemotron 3 Omni 30B A3B}
&43.12
&39.39
&\underline{7.07}
&3.41\\

Ministral 8B 2512
&36.95
&28.28
&17.17
&1.68\\

Ministral 3B 2512
&25.17
&20.20
&10.10
&3.15\\

Llama 3.3 70B Instruct
&39.48
&29.29
&27.27
&2.29\\

Llama 3.2 3B Instruct
&31.06
&20.20
&24.24
&5.14\\

IBM Granite 4.0 H Micro
&32.46
&26.67
&14.29
&2.38\\

Llama 3.2 1B Instruct
&27.78
&23.23
&9.09
&5.18\\

Microsoft Phi-4
&37.22
&31.43
&10.48
&1.81\\

\textbf{Gemma 3 4B IT}
&30.74
&22.22
&19.19
&\textbf{0.68}\\

\underline{Gemma 4 26B A4B IT}
&59.18
&\underline{54.55}
&10.10
&1.12\\

Gemma 4 31B IT
&59.01
&52.53
&14.14
&1.73\\

Gemma 3 12B IT
&38.38
&32.32
&23.23
&1.54\\

Nous Hermes 4 405B
&56.23
&40.40
&27.27
&3.13\\

Nous Hermes 3 Llama 3.1 405B
&60.86
&53.54
&14.14
&2.10\\

Amazon Nova Micro V1
&33.02
&16.19
&26.67
&2.40\\

Nous Hermes 4 70B
&40.90
&24.24
&29.29
&2.34\\

\bottomrule
\end{tabular} \\
OA = Overall Decision Accuracy. Configuration Range denotes the difference between the best and worst configuration-level accuracies. \textbf{Bold} indicates the best result, while \underline{underlined} values denote the second-best. Lower Range and SD indicate greater robustness.

\end{table}

\subsection{Robustness Across Configurations}
Table~\ref{tab:agentic} evaluates the robustness of the considered LLMs across different prompting strategies and decoding temperatures. xAI Grok 4.5 achieved the highest overall decision accuracy (63.89\%), followed by GPT-5.3 Chat (61.20\%), Claude Haiku 4.5 (60.95\%), Nous Hermes 3 Llama 3.1 405B (60.86\%), Gemma 4 26B A4B IT (59.18\%), Gemma 4 31B IT (59.01\%), and Qwen3.7-Max (58.42\%), demonstrating consistently strong performance across multiple evaluation settings. Furthermore, xAI Grok 4.5 also achieved the highest worst-case accuracy (56.19\%), indicating superior robustness even under the least favorable prompting and decoding configurations. Gemma 4 26B A4B IT achieved the second-highest worst-case accuracy (54.55\%), followed by GPT-5.3 Chat, Qwen3.7-Max, and Nous Hermes 3 Llama 3.1 405B (all 53.54\%). In terms of run-to-run consistency, Gemma 3 4B IT achieved the lowest mean run standard deviation (0.68), followed by Claude Haiku 4.5 (0.89), Gemma 4 26B A4B IT (1.12), Gemma 3 12B IT (1.54), Ministral 8B 2512 (1.68), and Gemma 4 31B IT (1.73), indicating highly consistent predictions across repeated runs. Overall, these results demonstrate that the strongest models combine high overall accuracy with strong worst-case performance, whereas robustness metrics such as configuration range alone are insufficient to characterize model quality, since models with consistently poor predictive performance may also exhibit relatively small configuration ranges.

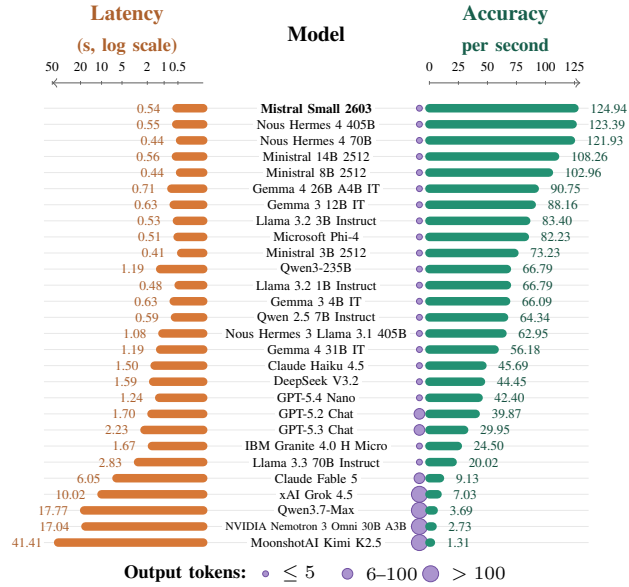
\begin{figure}[t]
\centering
\begingroup

\begin{tikzpicture}[x=0.55cm,y=0.50cm]
\def\leftbase{-2.72}
\def\rightbase{2.72}
\def\leftedge{-6.35}
\def\rightedge{6.35}
\def\barspan{3.63}
\def\tokenx{2.48}
\def\axisy{0.72}
\def\rowstart{0.05}
\def\rowstep{0.425}

\path[use as bounding box]
  (-7.60,2.85) rectangle (7.60,-12.72);

\newcommand{\efftightmodel}[1]{\resizebox{2.40cm}{!}{#1}}

\newcommand{\effrow}[5]{%
  \pgfmathsetmacro{\rowy}{\rowstart-\rowstep*(#1)}%
  \pgfmathsetmacro{\latencyx}
    {\leftbase-\barspan*ln(1+(#3)/0.4)/ln(126)}%
  \pgfmathsetmacro{\accx}
    {\rightbase+\barspan*(#4)/130}%
  \pgfmathsetmacro{\tokenradius}
    {ifthenelse((#5)<=5,1.1,ifthenelse((#5)<=100,2.0,3.0))}%

  \draw[black!9,line width=0.3pt]
    (-6.55,\rowy)--(6.55,\rowy);

  \draw[latencyorange,line width=3.2pt,line cap=round]
    (\latencyx,\rowy)--(\leftbase,\rowy);
  \node[anchor=east,font=\tiny,text=latencyorange!80!black]
    at ({\latencyx-0.16},\rowy) {#3};

  \node[
    anchor=center,
    font=\tiny,
    fill=white,
    inner xsep=1.2pt,
    inner ysep=0.2pt
  ] at (0,\rowy) {#2};

  \filldraw[
    fill=tokenpurple!65,
    draw=tokenpurple!90!black,
    line width=0.25pt
  ] (\tokenx,\rowy) circle[radius=\tokenradius pt];

  \draw[efficiencygreen,line width=3.2pt,line cap=round]
    (\rightbase,\rowy)--(\accx,\rowy);
  \node[anchor=west,font=\tiny,text=efficiencygreen!55!black]
    at ({\accx+0.16},\rowy) {#4};
}

\node[
  align=center,
  font=\footnotesize\bfseries,
  text=latencyorange!80!black
] at (-4.55,2.12)
  {Latency\\[0.75ex]{\scriptsize (s, log scale)}};

\node[font=\footnotesize\bfseries] at (0,2.02) {Model};

\node[
  align=center,
  font=\footnotesize\bfseries,
  text=efficiencygreen!65!black
] at (4.55,2.12)
  {Accuracy\\[0.75ex]{\scriptsize per second}};

\draw[->,black!60,line width=0.4pt]
  (\leftbase,\axisy)--(\leftedge,\axisy);
\foreach \latencytick in {0.5,1,2,5,10,20,50}{
  \pgfmathsetmacro{\latencytickx}
    {\leftbase-\barspan*ln(1+\latencytick/0.4)/ln(126)}
  \draw[black!60,line width=0.35pt]
    (\latencytickx,{\axisy-0.05})--(\latencytickx,{\axisy+0.05});
  \node[anchor=south,font=\tiny]
    at (\latencytickx,{\axisy+0.07}) {\latencytick};
}

\draw[->,black!60,line width=0.4pt]
  (\rightbase,\axisy)--(\rightedge,\axisy);
\foreach \acctick in {0,25,50,75,100,125}{
  \pgfmathsetmacro{\acctickx}
    {\rightbase+\barspan*\acctick/130}
  \draw[black!60,line width=0.35pt]
    (\acctickx,{\axisy-0.05})--(\acctickx,{\axisy+0.05});
  \node[anchor=south,font=\tiny]
    at (\acctickx,{\axisy+0.07}) {\acctick};
}

\effrow{0}{\textbf{Mistral Small 2603}}{0.54}{124.94}{2.00}
\effrow{1}{Nous Hermes 4 405B}{0.55}{123.39}{2.00}
\effrow{2}{Nous Hermes 4 70B}{0.44}{121.93}{2.00}
\effrow{3}{Ministral 14B 2512}{0.56}{108.26}{2.00}
\effrow{4}{Ministral 8B 2512}{0.44}{102.96}{2.00}
\effrow{5}{Gemma 4 26B A4B IT}{0.71}{90.75}{2.00}
\effrow{6}{Gemma 3 12B IT}{0.63}{88.16}{2.00}
\effrow{7}{Llama 3.2 3B Instruct}{0.53}{83.40}{2.00}
\effrow{8}{Microsoft Phi-4}{0.51}{82.23}{2.00}
\effrow{9}{Ministral 3B 2512}{0.41}{73.23}{2.00}
\effrow{10}{Qwen3-235B}{1.19}{66.79}{1.76}
\effrow{11}{Llama 3.2 1B Instruct}{0.48}{66.79}{2.00}
\effrow{12}{Gemma 3 4B IT}{0.63}{66.09}{2.00}
\effrow{13}{Qwen 2.5 7B Instruct}{0.59}{64.34}{2.00}
\effrow{14}{\efftightmodel{Nous Hermes 3 Llama 3.1 405B}}{1.08}{62.95}{1.00}
\effrow{15}{Gemma 4 31B IT}{1.19}{56.18}{2.00}
\effrow{16}{Claude Haiku 4.5}{1.50}{45.69}{4.00}
\effrow{17}{DeepSeek V3.2}{1.59}{44.45}{1.99}
\effrow{18}{GPT-5.4 Nano}{1.24}{42.40}{5.00}
\effrow{19}{GPT-5.2 Chat}{1.70}{39.87}{27.62}
\effrow{20}{GPT-5.3 Chat}{2.23}{29.95}{58.84}
\effrow{21}{IBM Granite 4.0 H Micro}{1.67}{24.50}{2.00}
\effrow{22}{Llama 3.3 70B Instruct}{2.83}{20.02}{2.00}
\effrow{23}{Claude Fable 5}{6.05}{9.13}{43.49}
\effrow{24}{xAI Grok 4.5}{10.02}{7.03}{512.43}
\effrow{25}{Qwen3.7-Max}{17.77}{3.69}{839.47}
\effrow{26}{\efftightmodel{NVIDIA Nemotron 3 Omni 30B A3B}}{17.04}{2.73}{1289.85}
\effrow{27}{MoonshotAI Kimi K2.5}{41.41}{1.31}{2532.51}

\node[anchor=east,font=\scriptsize\bfseries]
  at (-1.45,-12.25) {Output tokens:};

\filldraw[fill=tokenpurple!65,draw=tokenpurple!90!black,line width=0.3pt]
  (-1.22,-12.25) circle[radius=1.1pt];
\node[anchor=west,font=\scriptsize]
  at (-1.05,-12.25) {$\leq 5$};

\filldraw[fill=tokenpurple!65,draw=tokenpurple!90!black,line width=0.3pt]
  (0.75,-12.25) circle[radius=2pt];
\node[anchor=west,font=\scriptsize]
  at (0.97,-12.25) {$6$--$100$};

\filldraw[fill=tokenpurple!65,draw=tokenpurple!90!black,line width=0.3pt]
  (2.75,-12.25) circle[radius=3pt];
\node[anchor=west,font=\scriptsize]
  at (3.02,-12.25) {$>100$};

\end{tikzpicture}
\endgroup

\caption{Accuracy–latency trade-offs for model–provider configurations selected on the development set.}
\label{fig:Model_efficiency}
\end{figure}

\subsection{\textcolor{black}{Development-Stage Efficiency and Reliability}}

Figure~\ref{fig:Model_efficiency} compares the computational efficiency and output reliability of the evaluated LLMs using their best-performing configurations. To jointly assess predictive performance and inference efficiency, we report the Accuracy/Second metric, defined as the benchmark accuracy divided by the average inference latency. Among all models, Mistral Small 2603 achieved the highest efficiency with an Accuracy/Second score of 124.94, closely followed by Nous Hermes 4 405B (123.39), Nous Hermes 4 70B (121.93), Ministral 14B 2512 (108.26), and Ministral 8B 2512 (102.96), demonstrating that several medium-scale open-weight models provide an excellent balance between predictive performance and inference speed. Ministral 3B 2512 recorded the lowest inference latency (0.41~s), followed by Ministral 8B 2512 and Nous Hermes 4 70B (0.44~s), while Nous Hermes 3 Llama 3.1 405B produced the shortest outputs, averaging only one output token per benchmark instance. In contrast, frontier-scale reasoning models such as Qwen3.7-Max exhibited substantially higher latency (17.77~s) and considerably longer outputs (839.47 tokens), highlighting the computational cost associated with large-scale reasoning. Most evaluated models achieved a parse success rate of 100\%, demonstrating reliable adherence to the required output format, whereas GPT-5.2 Chat (98.99\%) and GPT-5.3 Chat (96.97\%) exhibited only minor parsing failures. By contrast, the NVIDIA Nemotron 3 family showed substantially lower parse success rates, ranging from 14.14\% to 61.62\%, which largely explains their poor benchmark performance despite their large parameter counts. Overall, these results demonstrate that high benchmark accuracy does not necessarily imply high computational efficiency. Instead, several compact and medium-scale open-weight models achieve a substantially better trade-off between predictive performance, inference latency, and output reliability than considerably larger frontier models.


\begin{table*}[t]
\centering
\color{black}
\caption{Held-Out Performance After Model-Specific Configuration Selection on the Development Set.}
\label{tab:heldout_generalization}
\scriptsize
\setlength{\tabcolsep}{1pt}
\renewcommand{\arraystretch}{1}
\begin{tabular}{lccccc}
\toprule
\textbf{Model}
& \textbf{Selected Configuration}
& \textbf{Development Acc. (\%)}
& \textbf{Held-Out Acc. (\%)}
& \textbf{95\% CI}
& \textbf{Gap (pp)} \\
\midrule

GPT-5.3 Chat
& 3-shot ($T=0.2$)
& $66.67 \pm 3.03$
& $\boldsymbol{52.00 \pm 0.72}$
& $[47.87,\,56.40]$
& $14.67$ \\

GPT-5.2 Chat
& 5-shot ($T=0.0$)
& $67.68 \pm 1.75$
& $\underline{49.40 \pm 1.44}$
& $[45.40,\,53.47]$
& $18.28$ \\

xAI Grok 4.5
& 3-shot ($T=0.5$)
& $\boldsymbol{70.48 \pm 3.30}$
& $49.07 \pm 0.81$
& $[45.07,\,53.33]$
& $21.41$ \\

Qwen3.7-Plus
& 3-shot ($T=0.8$)
& $67.68 \pm 3.50$
& $47.73 \pm 0.50$
& $[43.67,\,51.80]$
& $19.95$ \\

Qwen3.7-Max
& 3-shot ($T=0.2$)
& $65.66 \pm 1.75$
& $46.40 \pm 0.40$
& $[42.13,\,50.93]$
& $19.26$ \\

DeepSeek V3.2
& 3-shot ($T=0.0$)
& $\boldsymbol{70.48 \pm 1.65}$
& $46.07 \pm 0.50$
& $[41.93,\,50.40]$
& $24.41$ \\

Nous Hermes 4 70B
& 5-shot ($T=0.0$)
& $53.54 \pm 1.75$
& $45.87 \pm 0.61$
& $[41.87,\,50.07]$
& $7.67$ \\

Ministral 14B 2512
& 5-shot ($T=0.8$)
& $60.61 \pm 3.03$
& $45.27 \pm 0.50$
& $[41.13,\,49.60]$
& $15.34$ \\

Claude Haiku 4.5
& 5-shot ($T=0.0$)
& $\underline{68.57 \pm 0.00}$
& $44.80 \pm 0.00$
& $[40.40,\,49.20]$
& $23.77$ \\

Gemma 4 31B IT
& 5-shot ($T=0.5$)
& $66.67 \pm 0.00$
& $44.67 \pm 0.23$
& $[40.47,\,49.07]$
& $22.00$ \\

Gemma 4 26B A4B IT
& 3-shot ($T=0.2$)
& $64.65 \pm 1.75$
& $43.87 \pm 1.33$
& $[39.80,\,47.93]$
& $20.78$ \\

Nous Hermes 3 Llama 3.1 405B
& 5-shot ($T=0.2$)
& $67.68 \pm 1.75$
& $43.80 \pm 1.31$
& $[39.60,\,48.07]$
& $23.88$ \\

Nous Hermes 4 405B
& 5-shot ($T=0.0$)
& $67.68 \pm 1.75$
& $43.67 \pm 0.31$
& $[39.33,\,48.07]$
& $24.01$ \\

Gemma 3 12B IT
& 3-shot ($T=0.5$)
& $55.56 \pm 3.50$
& $42.47 \pm 0.23$
& $[38.27,\,46.80]$
& $13.09$ \\

Kimi K2.5
& 3-shot ($T=0.2$)
& $54.29 \pm 4.95$
& $41.87 \pm 0.12$
& $[37.73,\,46.13]$
& $12.42$ \\

Ministral 8B 2512
& 5-shot ($T=0.0$)
& $45.45 \pm 0.00$
& $41.40 \pm 0.00$
& $[37.20,\,45.80]$
& $4.05$ \\

Mistral Small 2603
& 5-shot ($T=0.0$)
& $67.09 \pm 1.71$
& $40.80 \pm 0.20$
& $[36.60,\,45.00]$
& $26.29$ \\

Qwen3-235B
& 3-shot ($T=0.8$)
& $63.64 \pm 1.75$
& $40.60 \pm 0.80$
& $[36.73,\,44.47]$
& $23.04$ \\

IBM Granite 4.0 H-Micro
& 5-shot ($T=0.2$)
& $40.95 \pm 1.65$
& $38.67 \pm 0.23$
& $[34.73,\,42.80]$
& $\boldsymbol{2.28}$ \\

Microsoft Phi-4
& 0-shot ($T=0.5$)
& $41.90 \pm 4.36$
& $38.33 \pm 0.58$
& $[34.60,\,42.20]$
& $\underline{3.57}$ \\

Llama 3.3 70B Instruct
& 5-shot ($T=0.8$)
& $56.57 \pm 1.75$
& $37.53 \pm 1.22$
& $[33.47,\,41.73]$
& $19.04$ \\

Llama 3.2 3B Instruct
& 5-shot ($T=0.0$)
& $44.44 \pm 1.75$
& $37.33 \pm 0.31$
& $[33.47,\,41.27]$
& $7.11$ \\

GPT-5.4 Nano
& 5-shot ($T=0.0$)
& $52.53 \pm 9.26$
& $37.33 \pm 0.31$
& $[33.73,\,40.93]$
& $15.20$ \\

Claude Fable 5
& 3-shot ($T=0.2$)
& $55.24 \pm 1.65$
& $36.07 \pm 0.46$
& $[32.07,\,40.20]$
& $19.17$ \\

Amazon Nova Micro V1
& 5-shot ($T=0.0$)
& $42.86 \pm 0.00$
& $34.40 \pm 1.20$
& $[30.33,\,38.60]$
& $8.46$ \\

Gemma 3 4B IT
& 5-shot ($T=0.5$)
& $41.41 \pm 1.75$
& $33.67 \pm 1.27$
& $[29.67,\,37.73]$
& $7.74$ \\

Qwen 2.5 7B Instruct
& 3-shot ($T=0.8$)
& $38.10 \pm 1.65$
& $29.20 \pm 0.35$
& $[25.33,\,33.20]$
& $8.90$ \\

NVIDIA Nemotron 3 Nano Omni 30B
& 5-shot ($T=0.5$)
& $46.46 \pm 3.50$
& $27.87 \pm 21.36$
& $[25.33,\,30.60]$
& $18.59$ \\

Llama 3.2 1B Instruct
& 3-shot ($T=0.8$)
& $32.32 \pm 14.00$
& $27.40 \pm 1.31$
& $[25.00,\,29.80]$
& $4.92$ \\

Ministral 3B 2512
& 0-shot ($T=0.0$)
& $30.30 \pm 3.03$
& $25.40 \pm 0.00$
& $[21.80,\,29.20]$
& $4.90$ \\

\bottomrule
\end{tabular}

\vspace{1mm}
\begin{minipage}{\textwidth}
\footnotesize
\textit{Note:} Models are ordered by decreasing held-out accuracy.
Development and held-out accuracies are reported as the mean and
sample standard deviation across three prespecified runs. Confidence
intervals are obtained by bootstrap resampling complete source
episodes, thereby preserving within-episode dependence. The
generalization gap is defined as
$\mathrm{Acc}_{\mathrm{dev}}-
\mathrm{Acc}_{\mathrm{held\text{-}out}}$.
Bold and underlined values denote the best and second-best results,
respectively. Lower values are preferable for the generalization gap,
and tied best values receive equal emphasis.
\end{minipage}
\end{table*}

\subsection{Held-Out Generalization}

\textcolor{black}{
Table~\ref{tab:heldout_generalization} evaluates generalization after each model's prompting strategy and decoding temperature were selected exclusively on the 35-instance human-verified development set and then frozen before evaluation on 500 episode-disjoint held-out decision instances. GPT-5.3 Chat achieves the highest held-out accuracy at $52.00\pm0.72\%$, followed by GPT-5.2 Chat at $49.40\pm1.44\%$ and xAI Grok 4.5 at $49.07\pm0.81\%$. The ranking differs substantially from that observed during development: although xAI Grok 4.5 and DeepSeek V3.2 jointly obtain the highest development accuracy of $70.48\%$, their held-out accuracies decrease to $49.07\%$ and $46.07\%$, respectively. Positive generalization gaps appear for every model and range from $2.28$ to $26.29$ percentage points, showing that selecting configurations and reporting performance on the same small development set would yield systematically optimistic estimates. IBM Granite 4.0 H-Micro exhibits the smallest gap, but its lower absolute accuracy shows that stability and predictive quality must be interpreted jointly. These results support the two-stage protocol by showing that performance on unseen source episodes provides a more rigorous and discriminative estimate of Physical AI decision-making capability than development-set performance alone.}

\subsection{Limitations}

Although \textit{PhysAI-Bench} provides a scalable and reproducible benchmark for evaluating agentic decision-making in Physical AI, several limitations remain. First, the current benchmark is instantiated from autonomous UAV missions. While the proposed benchmark construction methodology is general and can be extended to other Physical AI applications, including robotics, autonomous driving, embodied assistants, and industrial automation, the present benchmark primarily reflects the operational characteristics of autonomous UAV systems. Second, \textit{PhysAI-Bench} represents the decision context using structured textual descriptions derived from autonomous execution traces. Compared with decision-making based on rich multimodal observations, such as images, video, LiDAR, audio, and other sensor modalities, this formulation provides a simpler evaluation setting by abstracting low-level perception and focusing on high-level autonomous reasoning and action selection. Although modern foundation models ultimately represent these heterogeneous modalities as tokens, they introduce additional perception and representation challenges beyond the scope of the current benchmark. Extending \textit{PhysAI-Bench} to directly incorporate multimodal observations and evaluate vision--language models (VLMs) and multimodal large language models (MLLMs) therefore constitutes an important direction for future research.

\section{Conclusion}
\label{sec:conc}

In this paper, we introduced \textit{PhysAI-Bench}, a benchmark containing \(10{,}178\) multiple-choice decision instances derived from autonomous UAV mission traces. Each instance retains the operational information available at decision time, including mission constraints, physical conditions, sensor observations, MCP tool invocations, A2A interactions, and AI-native 6G network states, while excluding subsequent trace events. This formulation isolates context-aware action selection from broader perception and end-to-end task-completion capabilities. We evaluated 29 frontier LLMs using a two-stage protocol that separated configuration selection from held-out performance estimation. On the fixed, episode-disjoint set of \(500\) instances, GPT-5.3 Chat achieved the highest accuracy of \(52.00\%\), followed by GPT-5.2 Chat with \(49.40\%\) and Grok~4.5 with \(49.07\%\). Development-stage results showed that few-shot prompting generally improved performance, whereas decoding temperature had a limited effect. These findings confirm that reliable agentic decision-making remains challenging for current foundation models. Future work will extend the benchmark to robotics, autonomous driving, embodied AI, and other cyber-physical domains while incorporating richer multimodal observations, more diverse environments, and broader multi-agent scenarios.

\bibliographystyle{IEEEtran}
\bibliography{bibliography.bib}

\end{document}